%% file: Offline.tex
\documentclass{article}

\PassOptionsToPackage{numbers, compress}{natbib}
\usepackage[final]{neurips_2021}

\usepackage[utf8]{inputenc} % allow utf-8 input
\usepackage[T1]{fontenc}    % use 8-bit T1 fonts
\usepackage{hyperref}       % hyperlinks
\usepackage{url}            % simple URL typesetting
\usepackage{booktabs}       % professional-quality tables
\usepackage{amsfonts}       % blackboard math symbols
\usepackage{nicefrac}       % compact symbols for 1/2, etc.
\usepackage{microtype}      % microtypography
\usepackage{xcolor}         % colors

\usepackage{amsmath,amsthm,amssymb}
\usepackage{thmtools}
\usepackage{algorithm}
\usepackage[noend]{algpseudocode}
\usepackage{enumitem}
\usepackage{graphicx}
\usepackage{wrapfig}
\usepackage{subcaption}
\usepackage[all]{xy}
\usepackage{cancel}
\usepackage[figuresright]{rotating}

\definecolor{airforceblue}{rgb}{0.36, 0.54, 0.66}
\definecolor{lightgreen}{rgb}{0.56, 0.93, 0.56}
\definecolor{lightseagreen}{rgb}{0.13, 0.7, 0.67}
\definecolor{limegreen}{rgb}{0.2, 0.8, 0.2}
\definecolor{lincolngreen}{rgb}{0.11, 0.35, 0.02}
\definecolor{mediumseagreen}{rgb}{0.24, 0.7, 0.44}
\definecolor{napiergreen}{rgb}{0.16, 0.5, 0.0}
\definecolor{parisgreen}{rgb}{0.31, 0.78, 0.47}
\definecolor{teagreen}{rgb}{0.82, 0.94, 0.75}
\definecolor{yellow-green}{rgb}{0.6, 0.8, 0.2}
\definecolor{applegreen}{rgb}{0.55, 0.71, 0.0}

\newcommand*{\imgintext}[1]{%
  \raisebox{-.3\baselineskip}{%
    \includegraphics[
      height=\baselineskip,
      width=\baselineskip,
      keepaspectratio,
    ]{#1}%
  }%
}

\input{math_commands}

\title{Offline Reinforcement Learning with Reverse Model-based Imagination}

\author{%
	Jianhao Wang\thanks{Equal contribution.}~~$^1$, Wenzhe Li\footnotemark[1]~~$^1$, Haozhe Jiang$^1$, Guangxiang Zhu$^2$\thanks{Work done while Guangxiang was a Ph.D. student at Tsinghua University.}~~, Siyuan Li$^1$, \\ \textbf{Chongjie Zhang}$^1$ \\
	$^1$Institute for Interdisciplinary Information Sciences, Tsinghua University, China \\
	$^2$Baidu Inc., China \\
	\texttt{\{wjh19, lwz21, jianghz20\}@mails.tsinghua.edu.cn}\\
	\texttt{guangxiangzhu@outlook.com}\\
	\texttt{sy-li17@mails.tsinghua.edu.cn} \\
	\texttt{chongjie@tsinghua.edu.cn} \\
}

\begin{document}
	
	\maketitle

	\input{0-Abstract}
	\input{1-Introduction}

	\input{2-Preliminaries}
	\input{3-Methodology}
	\input{4-Experiments}
	\input{5-Related-Work}

	\input{6-Conclusion}

	\begin{ack}
    The authors would like to thank the anonymous reviewers, Zhizhou Ren, Kun Xu, and Hang Su for valuable and insightful  discussions and helpful suggestions.  This work is supported in part by Science and Technology Innovation 2030 – “New Generation Artificial Intelligence” Major Project (No. 2018AAA0100904), a grant from the Institute of Guo Qiang, Tsinghua University, and a grant from Turing AI Institute of Nanjing. 
    \end{ack}
	
	%\cite{kumar2019stabilizing}
	
    \bibliography{references}
    \bibliographystyle{unsrt}
    
    \newpage

	\newpage
	\input{7-Appendix}

	%\section{Appendix}
	
	%Optionally include extra information (complete proofs, additional experiments and plots) in the appendix.
	%This section will often be part of the supplemental material.
	
\end{document}

%% file: math_commands.tex
\usepackage{amsmath,amsfonts,bm}

\def\eqref#1{equation~\ref{#1}}
\def\1{\bm{1}}

\DeclareMathAlphabet{\mathsfit}{\encodingdefault}{\sfdefault}{m}{sl}
\SetMathAlphabet{\mathsfit}{bold}{\encodingdefault}{\sfdefault}{bx}{n}

\def\gA{{\mathcal{A}}}

\def\gD{{\mathcal{D}}}

\def\gJ{{\mathcal{J}}}

\def\gL{{\mathcal{L}}}
\def\gM{{\mathcal{M}}}
\def\gN{{\mathcal{N}}}

\def\gS{{\mathcal{S}}}

\def\sD{{\mathbb{D}}}
\newcommand{\E}{\mathbb{E}}

\newcommand{\Denv}{\gD_{\rm{env}}}
\newtheorem{definition}{Definition}

%% file: 0-Abstract.tex
\begin{abstract}
	In offline reinforcement learning (offline RL), one of the main challenges is to deal with the distributional shift between the learning policy and the given dataset. To address this problem,  recent offline RL methods attempt to introduce conservatism bias to encourage learning in high-confidence areas. Model-free approaches directly encode such bias into policy or value function learning using conservative regularizations or special network structures, but their constrained policy search limits the generalization beyond the offline dataset. Model-based approaches learn forward dynamics models with conservatism quantifications and then generate imaginary trajectories to extend the offline datasets. However, due to limited samples in offline datasets, conservatism quantifications often suffer from overgeneralization in out-of-support regions. The unreliable conservative measures will mislead forward model-based imaginations to undesired areas, leading to overaggressive behaviors. To encourage more conservatism, we propose a novel model-based offline RL framework, called \textit{\textbf{R}everse \textbf{O}ffline \textbf{M}odel-based \textbf{I}magination} (ROMI). We learn a reverse dynamics model in conjunction with a novel reverse policy,  which can generate rollouts leading to the target goal states within the offline dataset. These reverse imaginations provide informed data augmentation for model-free policy learning and enable conservative generalization beyond the offline dataset. ROMI can effectively combine with off-the-shelf model-free algorithms to enable model-based generalization with proper conservatism. Empirical results show that our method can generate more conservative behaviors and achieve state-of-the-art performance on offline RL benchmark tasks.

\end{abstract}

%% file: 1-Introduction.tex
\section{Introduction}
\label{sec:intro}

Deep reinforcement learning (RL) has achieved tremendous successes in a range of domains~\cite{silver2016mastering,mnih2015human,lillicrap2015continuous} by utilizing a large number of interactions with the environment. However, in many real-world applications, collecting sufficient exploratory interactions is usually impractical, because online data collection can be costly or even dangerous, such as in healthcare ~\cite{gottesman2019guidelines} and autonomous driving \cite{yu2018bdd100k}. To address this challenge, offline RL~\cite{lange2012batch,levine2020offline} develops a new learning paradigm that trains RL agents only with pre-collected offline datasets and thus can abstract away from the cost of online exploration ~\cite{fujimoto2019off,wu2019behavior,kumar2019stabilizing,kumar2020conservative,kidambi2020morel,yu2020mopo,lee2021representation,yu2021combo,agarwal2020optimistic,chen2020bail}. For such offline settings, recent studies demonstrate that directly applying the online RL algorithms can lead to poor  performance~\cite{fujimoto2019off,wu2019behavior,agarwal2020optimistic}. This phenomenon is primarily attributed to~\textit{distributional shift}~\cite{levine2020offline} between the learning policy and the behavior policy induced by the given dataset. With function approximation, the learning policy can overgeneralize the offline dataset and result in unexpected or dangerous behaviors. Thus developing techniques to handle distributional shift is becoming an active topic in the community of offline RL.

Recently, a variety of advanced offline RL algorithms have been proposed, which introduce conservatism bias and constrain the policy search in high-confidence regions induced by the offline dataset. Model-free offline RL methods ~\cite{fujimoto2019off,wu2019behavior,kumar2019stabilizing,kumar2020conservative} explicitly encode such bias into policy or value functions by using conservative regularizations or specially designed network structures. These methods often effectively address distributional shift issues, but their constrained policy search can limit the generalization beyond the offline dataset. In contrast, model-based offline RL ~\cite{kidambi2020morel,yu2020mopo,lee2021representation,yu2021combo} adopts more aggressive approaches. They first learn a forward dynamics model from the offline dataset with conservatism quantifications and then generate imaginary trajectories on high confidence regions to extend the offline dataset. Specifically, these methods may use a model-uncertainty quantification~\cite{kidambi2020morel,yu2020mopo}, representation learning of a robust model~\cite{lee2021representation}, or a conservative estimation of value functions~\cite{yu2021combo} to ensure the confidence of model-based rollouts. However, because samples in offline datasets are limited, conservatism quantifications often suffer from overgeneralization, especially in out-of-support regions. Thus, these unreliable measures can overestimate some unknown states and mislead forward model-based imaginations to undesired areas, leading to radicalism. In this paper, we will investigate a new direction in the context of model-based offline RL and propose reverse model imagination that enables effective conservative generalization.

\begin{figure}
	\centering
	\vspace{-0.1in}
	\includegraphics[width=\linewidth]{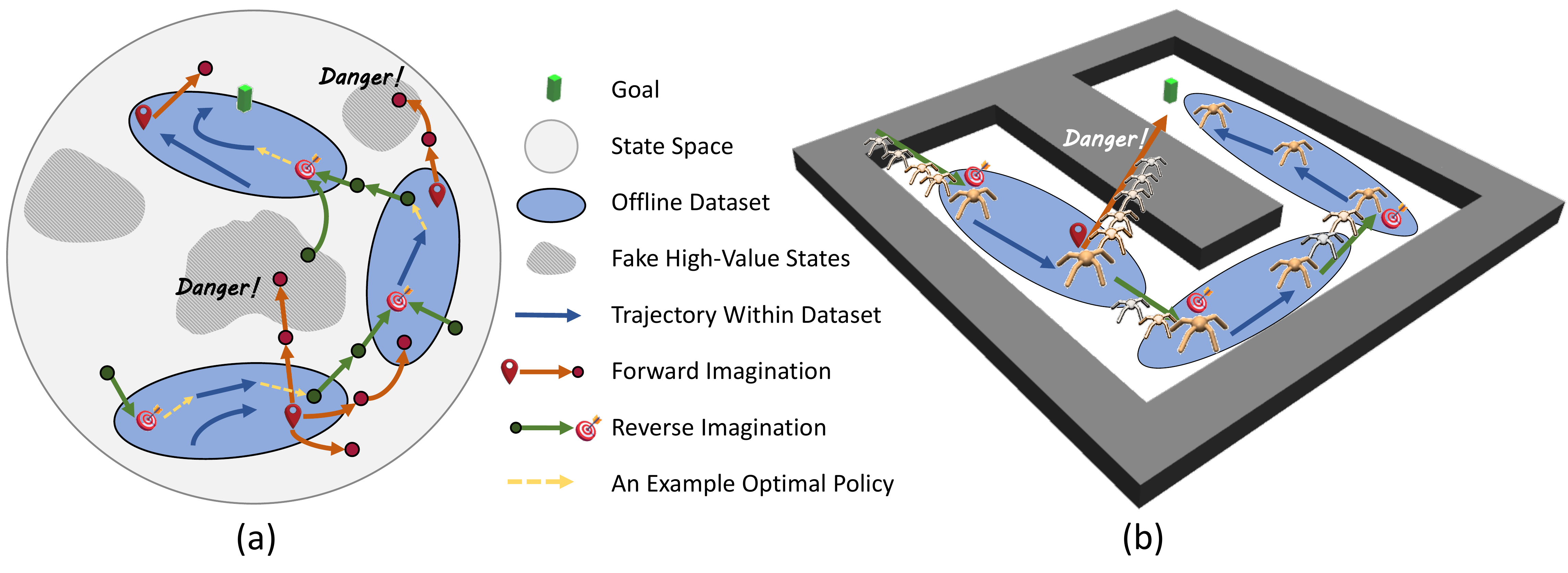}
	\caption{(a) Basic idea of ROMI. (b) A concrete RL task to demonstrate the superiority of ROMI.
	}
	\label{fig:motivation}
\end{figure}

We use Figure \ref{fig:motivation}a to illustrate our basic ideas. When an offline dataset contains expert or nearly optimal trajectories, model-free offline RL methods ~\cite{fujimoto2019off,wu2019behavior,kumar2019stabilizing,kumar2020conservative} show promising performance. % by explicitly imposing conservative bias. 
In other situations, such as in Figure \ref{fig:motivation}a, when the optimal policy requires a composition of multiple trajectories (\imgintext{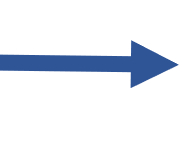}) in the offline dataset, model-free offline RL usually fails because it may get stuck in isolated regions of the offline dataset (\imgintext{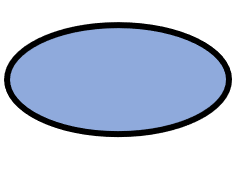}). In contrast, model-based approaches~\cite{kidambi2020morel,yu2020mopo,lee2021representation,yu2021combo} have advantages of connecting trajectories in the offline dataset by generating bridging rollouts. When using forward dynamics models, model-based methods can generate aggressive rollouts from the dataset to outside areas (\imgintext{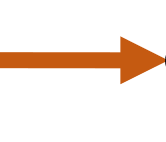}). Such {\color{brown}\textbf{\textit{forward imaginations}}} potentially discover a better policy outside the offline dataset, but may also lead to undesirable regions consisting of fake high-value states (\imgintext{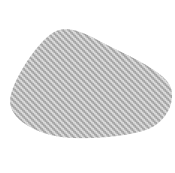}) due to overgeneralization errors. 
%The errors are usually very large for the out-of-support data (the outside of \imgintext{figures/x3.png}) because dynamics models are usually learned on the support of the offline dataset (\imgintext{figures/x3.png}). 
Things will be different if we reverse the model imagination. {\color{napiergreen}\textbf{\textit{Reverse imaginations}}} (\imgintext{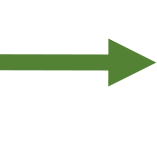}) generate possible traces leading to target goal states (\imgintext{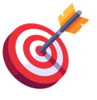}) inside the offline dataset, which provides a conservative way of augmenting the offline dataset. We assume that the reverse dynamics model has similar accuracy to the forward model. In this way, reverse models not only maintain the generalization ability to interpolate between given trajectories and potential better policies but also avoid radical model imagination by enabling bidirectional search in conjunction with the existing forward real trajectories (\imgintext{figures/x5.png}) in the offline dataset. The scenario illustrated in Figure \ref{fig:motivation}a is common in RL tasks~\cite{fu2020d4rl}, where the agent may encounter new obstacles or unexpected dangers at any time (e.g., bumping into walls like in Figure \ref{fig:motivation}b or falling). As many unexpected data are not recorded by an offline dataset, there are likely large generalization errors outside the support of the dataset. Thus, reverse models that can generate conservative imaginations are more appealing for real-world offline RL tasks.

Based on the above observation, we present a novel model-based offline RL framework, called \textit{\textbf{R}everse \textbf{O}ffline \textbf{M}odel-based \textbf{I}magination} (ROMI). ROMI trains a reverse dynamics model and generates backward imaginary trajectories by a backtracking rollout policy. This rollout policy is learned by a conditional generative model to produce diverse reverse actions for each state and lead to the unknown data space with high probabilities. ROMI ensures start states of the reverse rollout trajectories (i.e., target goals of forward trajectories) are within the offline dataset, and thus naturally imposes conservative constraints on imaginations. With this conservative data augmentation, ROMI has the advantage of effectively combining with off-the-shelf model-free algorithms (e.g., BCQ~\cite{fujimoto2019off} and CQL~\cite{kumar2020conservative}) to further strengthen its generalization with proper conservatism. Taken together, the whole ROMI framework enables extensive interpolation of the dataset and potentially better performance (contributed by diverse policy) in a safe manner (contributed by reverse imagination). To our best knowledge, ROMI is the first offline RL approach that utilizes reverse model-based imaginations to induce conservatism bias with data augmentation. It provides a novel bidirectional learning paradigm for offline RL, which connects reverse imaginary trajectories with pre-collected forward trajectories in the offline dataset. Such a bidirectional learning paradigm shares similar motivations with humans' bidirectional reasoning. Studies from Psychology \cite{holyoak1999bidirectional} show that, during the decision-making process, humans not only consider the consequences of possible future actions from a forward view but also imagine possible traces leading to the ideal goal through backward inference.

%ROMI takes advantages of static rollout policies. It is agnostic to policy learning algorithms and can be used as a plug-and-play module for any model-free offline RL method. 
We conduct extensive evaluations on the D4RL offline benchmark suite~\cite{fu2020d4rl}. Empirical results show that ROMI significantly outperforms state-of-the-art model-free and model-based baselines. Our method achieves the best or comparable performance on 16 out of 24 tasks among all algorithms. Ablation studies verify that the reverse model imagination can effectively generate more conservative behaviors than forward model imagination. Videos of the experiments are available online\footnote{\url{https://sites.google.com/view/romi-offlinerl/}.}.

%% file: 2-Preliminaries.tex
\section{Preliminaries}
\label{sec:preliminaries}
%\wl{follow the background section in backtrace model paper}

We consider a Markov decision process (MDP) defined by a tuple $\gM=(\gS,\gA,T,r,\mu_0,\gamma)$, where $\gS$ and $\gA$ denote the state space and the action space, respectively. $T(s'\vert s,a):\gS\times\gA\times\gS\rightarrow\mathbb{R}$ denotes the transition distribution function, $r(s,a):\gS\times\gA\rightarrow\mathbb{R}$ denotes the reward function, $\mu_0:\gS\rightarrow[0, 1]$ is the initial state distribution, and $\gamma\in(0,1)$ is the discount factor. Moreover, we denote the reverse transition distribution by $T_r(s\vert s',a)=T^{-1}:\gS\times\gA\times\gS\rightarrow\mathbb{R}$. The goal of an RL agent is to optimize a policy $\pi(a\vert s):\gS\times\gA\rightarrow\mathbb{R}$ that maximizes the expected cumulative reward, i.e., $\gJ(\pi)=\E_{s_0\sim\mu_0,s_{t+1}\sim T(\cdot{|s_t,\pi(s_t)})}\left[\sum_{t=0}^\infty\gamma^tr(s_t,\pi(s_t))\right]$. 

In the offline RL setting, the agent only has access to a static dataset $\Denv=\left\{(s,a,r,s')\right\}$ and is not allowed to interact with the environment for additional online explorations. The data can be collected through multi-source logging policies and we denote the empirical distribution of behavior policy in a given dataset $\Denv$ collected by $\pi_D$ . Logging policies are not accessible in our setting.

Model-based RL methods aim at performing planning or policy searches based on a learned model of the environment. They usually learn a dynamics model $\widehat{T}$ and a reward model $\widehat{r}(s, a)$ from a collection of environmental data in a self-supervised manner. 
Most of the existing approaches use the forward model $\widehat{T}_f(s'\vert s,a)$ for dynamics learning, but we will show in Section~\ref{sec:methodology} that the reverse dynamics model $\widehat{T}_r(s\vert s',a)$ can induce more conservatism and is critical for the offline RL tasks.

%$\widetilde{p}_\phi(s, r\vert s',a)$. 

%$\widehat{p}_\phi(s', r\vert s,a)$ 

%For simplicity, we denote the approximate dynamcis and reward model by $p_\phi$. 

%% file: 3-Methodology.tex
\section{Reverse Offline Model-based Imagination}
\label{sec:methodology}

In the offline RL setting, the agent can only access a given dataset without additional online exploration. Model-based offline RL algorithms usually face three challenges in this setting. (i) The offline dataset has limited samples and generalization over the given data is required. (ii) There are a lot of uncertainties that are difficult to estimate out of the support of the dataset. (iii) The model-based rollouts cannot receive feedback from the real environment. Thus it is important to augment diverse data and keep conservative generalization at the same time. In this section, we will introduce the \textit{\textbf{R}everse \textbf{O}ffline \textbf{M}odel-based \textbf{I}magination} (ROMI) framework to combine model-based imagination with model-free offline policy learning. This framework encourages diverse augmentation of model-based rollouts and enables conservative generalization of the generated imaginations.

We use a concrete example to show the superiority of our framework in Figure \ref{fig:motivation}b. The agent navigates in a u-maze to reach the goal (\imgintext{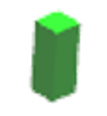}). We gather an offline dataset consisting of three data blocks (\imgintext{figures/x3.png}). This dataset only contains the trajectories that do not hit the wall, thus the agent will be unaware of the walls during the offline learning process. In this situation, the conventional forward model may generate a greedy and radical imagination (\imgintext{figures/x6.png}) that takes a shortcut to go straight to the goal (whose value function may be overestimated) and hit the wall. On the contrary, ROMI adopts backward imaginations and generates traces that can lead to the target goal states in the support of the dataset  (\imgintext{figures/x7.png}). Such imaginations can be further connected with the given forward trajectories (\imgintext{figures/x5.png}) in the dataset to form an optimal policy. In this example, we assume that the model error is mainly due to the prediction of the OOD states (outside of the support of the data), then we can expect reverse models to help combat overestimation since they prevent rollout trajectories that end in the OOD states. A more detailed illustrative example is deferred to Appendix \ref{appendix_example}. 

Specifically, our framework consists of two main components: (i) a reverse model learned from the offline dataset, (ii) a diverse rollout policy to generate reverse actions that are close to the dataset. ROMI pre-generates the model-based trajectories under the rollout policy based on the reverse model. Then we use the mixture of the imaginary data and the original dataset to train a model-free offline agent. The whole algorithm is shown in Algorithm \ref{alg:romi}.

\textbf{Training the reverse model.} To backtrace the dynamics and reward function of the environment, we introduce a reverse model to estimate the reverse dynamics model $\widehat{T}_r(s\vert s',a)$ and reward model $\widehat{r}(s, a)$ from the offline dataset simultaneously. For simplicity, we unify the dynamics and reward function into our reverse model $p(s, r\vert s',a)$, i.e., 
\begin{align}
p(s, r\vert s',a) = p(s\vert s',a) p(r\vert s',a,s) = T_r(s\vert s',a) p(r\vert s,a),
\end{align}
where we assume that the reward function only depends on the current state and action. This unified reverse model represents the probability of the current state and immediate reward conditioned on the next state and current action. We parameterize it by $\phi$  and optimize it by minimizing the loss function $\gL_M(\phi)$  (which is equivalent to maximizing the log-likelihood):
\begin{align}\label{eq:model_loss}
	\gL_M(\phi)=\mathop{\E}_{(s,a,r,s')\sim \Denv}\left[-\log \widehat{p}_\phi(s, r|s', a)\right],
\end{align}
where $\Denv$ is the offline dataset. 

\textbf{Training the reverse rollout policy.} To encourage diversity for the reverse model-based imagination near the dataset, we train a generative model $\widehat{G}_\theta(a|s')$, which samples diverse reverse actions from the offline dataset using stochastic inference. Specifically, we use a conditional variational auto-encoder (CVAE)~\cite{sohn2015learning,fujimoto2019off} to represent the diverse rollout policy $\widehat{G}_\theta(a|s')$, which is parameterized by $\theta$ and depends on the next state. The rollout policy $\widehat{G}_\theta(a|s')$ contains two modules: (i) an action encoder $\widehat{E}_\omega(s',a)$ that outputs a latent vector $z$ under the gaussian distribution $z\sim\widehat{E}_\omega(s',a)$, and (ii) an action decoder $\widehat{D}_\xi(s',z)$ whose input is the latent vector $z$ and reconstructs the given action $\widetilde{a}=\widehat{D}_\xi(s',z)$. The action encoder and decoder are parameterized by $\omega$ and $\xi$, respectively. For simplicity, denote the parameters of the rollout policy $\widehat{G}_\theta(a|s')$ by $\theta=\{\omega, \xi\}$.
We defer the detailed discussion of CVAE to Appendix \ref{appendix_CVAE}. 

We train the rollout policy $\widehat{G}_\theta(a|s')$ by maximizing the variational lower bound $\gL_p(\theta)$, 
%To optimize the variational lower bound of CVAE, we introduce a loss function of $\widehat{G}_\theta(s')$
%We defer the introdution of VAEs to Appendix. 
%$\widehat{G}_\theta(s')=\{\widehat{E}_\omega(s',a), \widehat{D}_\xi(s',z)\}$ $\theta=\{\omega, \xi\}$.
%$\widehat{E}_\omega(s',a)$ is a gaussian distribution $\gN(\mu,\sigma)$.
%We optimize the context encoder and the dynamics models by minimizing the following loss function:
\begin{align}\label{eq:rollout_policy_loss}
\gL_p(\theta)=\mathop{\E}_{(s,a,r,s')\sim \Denv,z\sim \widehat{E}_\omega(s',a)}\left[\left(a-\widehat{D}_\xi(s',z)\right)^2+D_{\text{KL}}\left(\widehat{E}_\omega(s',a)\|\gN(0,\bm{I})\right)\right],
\end{align}
where $\bm{I}$ is an identity matrix. To optimize such loss function, we adopt similar optimization techniques as those used in the context of image generalization and video prediction ~\cite{sohn2015learning,fujimoto2019off}.

After the rollout policy $\widehat{G}_\theta(a|s')$ is well trained, we can sample reverse actions $\widehat{a}$ based on the policy. We first draw a latent vector from the multivariate normal distribution, $\widehat{z}\sim\gN(0,\bm{I})$, and then utilize the action decoder to sample actions conditioned on the next state, $\widehat{a} = \widehat{D}_\xi(s',\widehat{z})$. To explore more possibilities, the rollout policy $\widehat{G}_\theta(a|s')$ uses stochastic layers to generate a variety of reverse actions for multiple times. In addition, if we need an extremely diverse policy in easy tasks, we can replace the generative model based policy with a uniformly random rollout policy.

%similar to BCQ~\cite{fujimoto2019off}
%reverse rollout policy $\widehat{G}_\theta(s')$.
%Sampling $\widehat{a} \sim \widehat{G}_\theta(s')$: 
%sample $\widehat{z}\sim\gN(0,\bm{I})$ and $\widehat{a} = \widehat{D}_\xi(s',\widehat{z})$. 

\textbf{Combination with model-free algorithms.} Based on the learned reverse dynamics model and the reverse rollout policy, ROMI can generate reverse imaginations. We collect these rollouts to form a  model-based buffer  $\gD_{\text{model}}$ and further compose the total dataset with the offline dataset, $\gD_{\text{total}} = \Denv \cup \gD_{\text{model}}$. 
Since our rollout policy is agnostic to policy learning, such buffer can be obtained before the policy learning stage, i.e., ROMI can be combined with any model-free offline RL algorithm (e.g., BCQ~\cite{fujimoto2019off} or CQL~\cite{kumar2020conservative}). Specifically, during the model-based imagination, we sample the target state $s_{t+1}$ from the dataset $\Denv$, and generate reverse imaginary trajectory $\widehat{\tau}$ with the rollout horizon $h$ by the reverse model $\widehat{p}_\phi$ and rollout policy $\widehat{G}_\theta$:
\begin{align*}
	\widehat{\tau}=\left\{(s_{t-i},a_{t-i},r_{t-i},s_{t+1-i})~\big|~a_{t-i} \sim \widehat{G}_\theta\left(\cdot|s_{t+1-i}\right) \text{ and } s_{t-i}, r_{t-i} \sim \widehat{p}_\phi\left(\cdot\vert s_{t+1-i}, a_{t-i}\right)\right\}_{i=0}^{h-1}.
\end{align*}
We gather trajectories $\widehat{\tau}$ to form the buffer $\gD_{\text{model}}$ and further combine it with $\Denv$ to obtain $\gD_{\text{total}}$. Then we will run the model-free offline policy learning algorithm on the total buffer to derive the final policy $\pi_{\text{out}}$. Compared to existing model-based approaches~\cite{kidambi2020morel,yu2020mopo,lee2021representation,yu2021combo}, ROMI provides informed data augmentation to extend the offline dataset. It is agnostic to policy optimization and thus can be regarded as an effective and flexible plug-in component to induce conservative model-based imaginations for offline RL.

\begin{algorithm}[htp]
	\caption{ROMI: Reverse Offline Model-based Imagination}\label{alg:romi}
	\begin{algorithmic}[1]
		\State {\bfseries Require:} Offline dataset $\Denv$, rollout horizon $h$, the number of iterations $C_\phi,C_\theta,T$, learning rates $\alpha_\phi,\alpha_\theta$, model-free offline RL algorithm (i.e., BCQ or CQL)
		\State Randomly initialize reverse model parameters $\phi$
		\For{$i=0\dots C_\phi-1$} \Comment{Learning a reverse dynamics model $\widehat{p}_\phi$}
		\State Compute $\gL_M$ using the dataset $\Denv$
		\State Update $\phi\gets\phi-\alpha_\phi\nabla_\phi\gL_M$
		\EndFor
		\State Randomly initialize rollout policy parameters $\theta$
		\For{$i=0\dots C_\theta-1$} \Comment{Learning a diverse rollout policy $\widehat{G}_\theta$}
		\State Compute $\gL_p$ using the dataset $\Denv$
		\State Update $\theta\gets\theta-\alpha_\theta\nabla_\theta\gL_p$
		\EndFor
		\State Initialize the replay buffer $\gD_{\text{model}}\gets\varnothing$
		\For{$i=0\dots T-1$} \Comment{Collecting the replay buffer $\gD_{\text{model}}$}
		\State Sample target state $s_{t+1}$ from the dataset $\Denv$
		\State Generate reverse model rollout $\widehat{\tau}=\left\{(s_{t-i},a_{t-i},r_{t-i},s_{t+1-i})\right\}_{i=0}^{h-1}$ from $s_{t+1}$ by drawing samples from the dynamics model $\widehat{p}_\phi$ and rollout policy $\widehat{G}_\theta$
		\State Add model rollouts to replay buffer,  $\gD_{\text{model}}\gets\gD_{\text{model}}\cup\left\{(s_{t-i},a_{t-i},r_{t-i},s_{t+1-i})\right\}_{i=0}^{h-1}$ %\Comment{Add model rollouts to $\mathcal{D}_{\text{model}}$}
		\EndFor
		\State Compose the final dataset $\gD_{\text{total}} \gets \Denv \cup \gD_{\text{model}}$
		\State Combine model-free offline RL algorithms to derive the final policy $\pi_{\text{out}}$ using the dataset $\gD_{\text{total}}$
		\State {\bfseries Return:} $\pi_{\text{out}}$
	\end{algorithmic}
\end{algorithm}

%% file: 4-Experiments.tex
\section{Experiments}
\label{sec:experiments}

In this section, we conduct a bunch of experiments in the offline RL benchmark~\cite{fu2020d4rl} to answer the following questions: 
%(1) Can the reverse model enjoy similar dynamic accuracy as the forward model (see Table~\ref{tab:modelerror})? 
(i) Does ROMI outperform the state-of-the-art offline RL baselines (see Table~\ref{tab:maze} and~\ref{tab:mujoco})?
%Does ROMI improve performance over its base model-free methods (see Table~\ref{tab:maze}~and~\ref{tab:mujoco})? 
(ii) Does ROMI achieve excellent performance because of the reverse model-based imagination (see Section~\ref{sec:ablation})? (iii) Is CVAE-based rollout policy critical for ROMI (see Table~\ref{tab:abla_rbc})? (iv) Compared with the forward imagination, does ROMI trigger more conservative and effective behaviors (see Figure~\ref{fig:visualization})?

\subsection{Evaluation Environments}
\label{sec:environments}

\begin{figure}
	\centering
	\vspace{-0.1in}
	\includegraphics[width=0.98\linewidth]{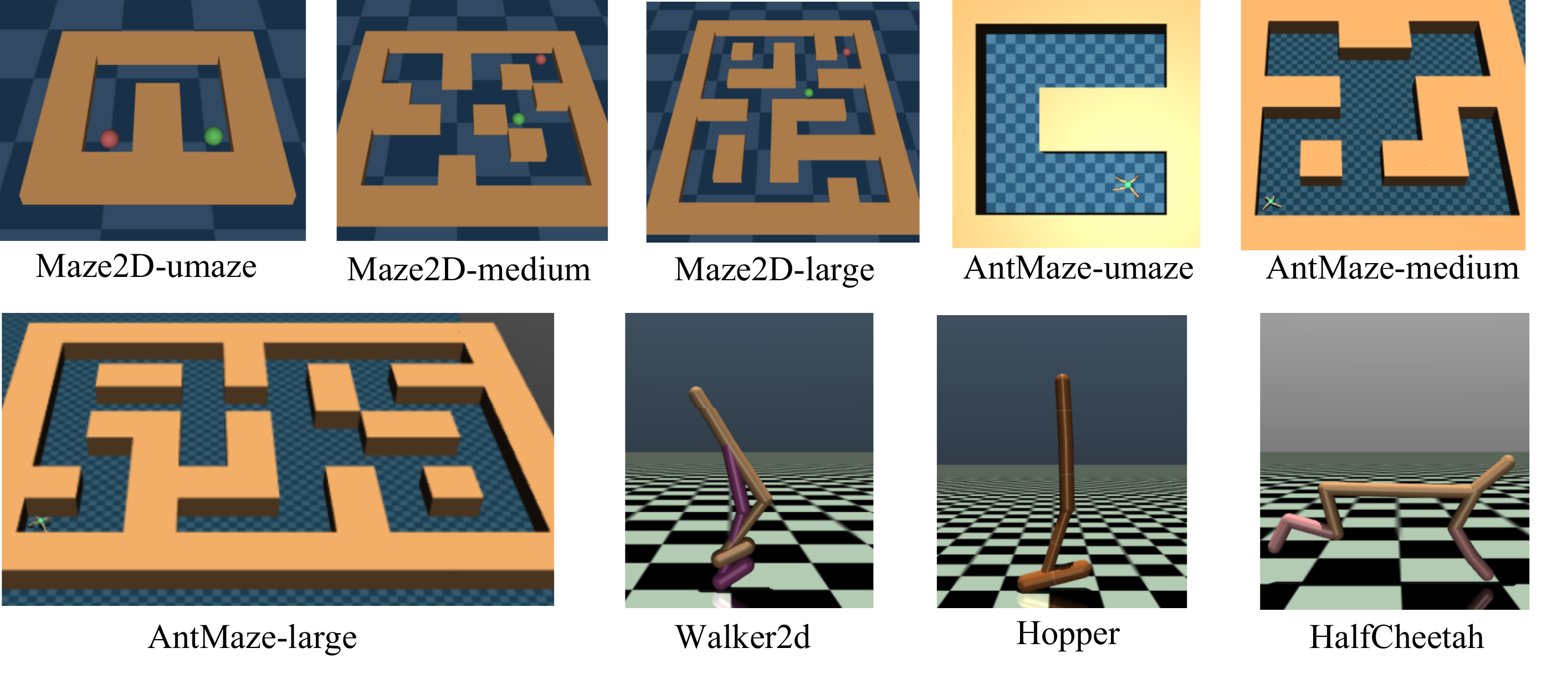}
	\caption{Experimental environments.
	}
	\label{fig:settings}
\end{figure}

We evaluate ROMI on a wide range of domains in the D4RL benchmark~\citep{fu2020d4rl}, including the Maze2D domain, the Gym-MuJoCo tasks, and the AntMaze domain. Figure~\ref{fig:settings} shows the snapshots of nine environments used in our experiments. We defer the quantification of ROMI's model accuracy in these domains to Appendix \ref{appendix_model_acc} and empirical evaluations show that reverse models have comparable accuracy, if not worse, than forward models.

\textbf{Maze2D.} The \textit{maze2d} domain requires a 2D agent to learn to navigate in the maze to reach a fixed target goal and stay there. As shown in Figure~\ref{fig:settings}, there are three maze layouts (i.e., \textit{umaze}, \textit{medium}, and \textit{large}) and two dataset types (i.e., \textit{sparse} and \textit{dense} reward singal) in this domain. The dataset of each layout is generated by a planner moving between randomly sampled waypoints. From the detailed discussion of the D4RL benchmark~\citep{fu2020d4rl}, we found that the agents in the \textit{maze2d} dataset are always moving on the clearing and will not stay in place. We will visualize the dataset of \textit{mazed2d-umaze} in Section~\ref{sec:visualization}.

%as shown in Figure~\ref{fig:settings}.
%\wl{I think we can directly use the figure in d4rl paper.}
%While forward models can stitch together parts of different trajectories, they are likely to overly generalize out of dataset and perform undesired behavior (e.g., try to go through the wall).

\textbf{Gym-MuJoCo.} The Gym-MuJoCo tasks consist of three different environments (i.e., \textit{walker2d}, \textit{hopper}, and \textit{halfcheetah}), and four types of datasets (i.e., \textit{random}, \textit{medium}, \textit{medium-replay}, and \textit{medium-expert}). \textit{Random} dataset contains experiences selected by a random policy. \textit{Medium} dataset contains experiences from an early-stopped SAC policy. \textit{Medium-replay} dataset records the samples in the replay buffer during the training of the "medium" SAC policy. \textit{Medium-expert} dataset is mixed with suboptimal samples and samples generated from an expert policy. 

\textbf{AntMaze.} The \textit{antmaze} domain combines challenges of the previous two domains. The policy needs to learn to control the robot and navigate to the goal simultaneously. This domain also contains three different layouts (i.e., \textit{umaze}, \textit{medium}, and \textit{large}) shown in Figure~\ref{fig:settings}. D4RL benchmark~\citep{fu2020d4rl} introduces three flavors of datasets (i.e., \textit{fixed}, \textit{diverse}, and \textit{play}) in this setting, which commands the ant from different types of start locations to various types of goals.
%Given the complex mechanisms of controlling robots, there is little chance that forward model imagination can generate trajectories to improve policy, as the robot may even fall down at the first step.

%summarizes the rollout error of both forward and reverse model under different domains.
\subsection{Overall Results}

In this subsection, the experimental results are presented in Table~\ref{tab:maze} and~\ref{tab:mujoco}, which are evaluated in the D4RL benchmark tasks~\citep{fu2020d4rl} illustrated in Figure~\ref{fig:settings}. We compare ROMI with 11 state-of-the-art baselines: \textit{MF} denotes the best performance from offline model-free algorithms, including BCQ~\cite{fujimoto2019off}, BEAR~\cite{kumar2019stabilizing}, BRAC-v, BRAC-p~\cite{wu2019behavior}, BAIL~\cite{chen2020bail}, and CQL~\cite{kumar2020conservative}; \textit{MB} denotes the best performance from offline model-based algorithms, including MOPO~\cite{yu2020mopo}, MOReL \citep{kidambi2020morel}, Repb-SDE~\cite{lee2021representation}, and COMBO \citep{yu2021combo}; \textit{BC} denotes the popular behavior cloning from the dataset in the imitation learning. The implementation details of ROMI and these algorithms are deferred to Appendix \ref{appendix_alg_details}. Towards fair evaluation, all experimental results are illustrated with the averaged performance with $\pm$ standard deviation over three random seeds.

We evaluate ROMI in nine D4RL benchmark domains with 24 tasks. Our experiments show that ROMI significantly outperforms baselines and achieves the best or comparable performance on 16 out of 24 continuous control tasks. Table~\ref{tab:maze} and~\ref{tab:mujoco} only contain the best performance of \textit{MF} and \textit{MB} categories\footnote{In Table~\ref{tab:mujoco}, each score of COMBO or MOReL is the better one between the score our reproduction and their reported score.}. We defer the pairwise comparison of ROMI and each baseline to Appendix \ref{appendix_full_exp}, which can demonstrate that ROMI in conjunction with off-the-shelf model-free methods (i.e., BCQ~\cite{fujimoto2019off} and CQL~\cite{kumar2020conservative}) can outperform all offline RL baselines. Specifically, we denote the methods with ROMI as ROMI-BCQ, and ROMI-CQL, respectively. The suffix \textit{-BCQ} or \textit{-CQL} indicates that ROMI adopts BCQ~\cite{fujimoto2019off} or CQL~\cite{kumar2020conservative} as base learning algorithms for policy optimization.

Table~\ref{tab:maze} shows that ROMI-BCQ is the best performer on 10 out of 12 tasks in the \textit{maze2d} and \textit{antmaze} domains. In these tasks, model-free methods can achieve reasonable performance, while current model-based algorithms with forward imagination cannot perform well, especially in \textit{antmaze}. This may be because that messy walls (see Figure~\ref{fig:settings}) are unknown from the given datasets, and the forward imagination may lead RL agent to bump into walls and lose the game. In contrast, reverse model-based imagination can avoid directing agents towards unconfident area, which makes a safe way to imagine in the complex domains. Table~\ref{tab:mujoco} shows that ROMI-CQL achieves the best performer on six out of 12 tasks in \textit{gym} domain. CQL is the best performer among all model-free methods, and ROMI-CQL can further outperform CQL on four challenging tasks. In this domain, current model-based methods perform pretty well. We argue that similar to forward imagination, reverse direction can also generalize beyond offline datasets for better performance in the relatively safe tasks, i.e., there is no obstacle around, as shown in Figure~\ref{fig:settings}.

\begin{table}
	\centering
	\caption{Performance of ROMI and best performance of prior methods on the \textit{maze} and \textit{antmaze} domains, on the normalized return metric proposed by D4RL benchmark~\citep{fu2020d4rl}. Scores roughly range from 0 to 100, where 0 corresponds to a random policy performance and 100 corresponds to an expert policy performance. \textit{med} is short for \textit{medium}.}
	\begin{tabular}{l|c|c|c|c}
		\toprule
		Environment & BC & ROMI-BCQ & MF & MB \\
		\midrule
		sparse-maze2d-umaze & -3.2 & \textbf{139.5}$~\pm~$3.6 & 65.7$~\pm~$6.9$^\text{BEAR}$ & 76.4$~\pm~$19.2$^\text{COMBO}$ \\
		sparse-maze2d-med & -0.5 & \textbf{82.4}$~\pm~$15.2 & 70.6$~\pm~$34.3$^\text{BRAC-v}$ & 68.5$~\pm~$83.6$^\text{COMBO}$ \\
		sparse-maze2d-large & -1.7 & \textbf{83.1}$~\pm~$22.1 & \textbf{81.0}$~\pm~$65.3$^\text{BEAR}$ & 14.1$~\pm~$10.7$^\text{COMBO}$ \\
		dense-maze2d-umaze & -6.9 & \textbf{98.3}$~\pm~$2.5 & 51.5$~\pm~$8.2$^\text{BRAC-p}$ & $\ $94.3$~\pm~$13.6$^\text{Repb-SDE}$ \\
		dense-maze2d-med & 2.7 & \textbf{102.6}$~\pm~$32.4 & 41.7$~\pm~$2.0$^\text{BAIL}$ & 84.2$~\pm~$9.5$^\text{COMBO}$ \\
		dense-maze2d-large & -0.3 & \textbf{124}$~\pm~$1.3& \textbf{133.0}$~\pm~$25.5$^\text{BEAR}$ & 36.8$~\pm~$12.4$^\text{MOPO}$ \\ \midrule
		fixed-antmaze-umaze & 82.0 & 68.7$~\pm~$2.7 & 75.3$~\pm~$13.7$^\text{BCQ}$ & \textbf{80.3}$~\pm~$18.5$^\text{COMBO}$ \\
		play-antmaze-med & 0.0 & \textbf{35.3}$~\pm~$1.3 & 1.7$~\pm~$1.0$^\text{BAIL}$ & 0.0 \\
		play-antmaze-large & 0.0 & \textbf{20.2}$~\pm~$14.8 & 2.2$~\pm~$1.3$^\text{BAIL}$ & 0.0 \\
		diverse-antmaze-umaze & 47.0 & \textbf{61.2}$~\pm~$3.3 & 54.0$~\pm~$15.0$^\text{BAIL}$ & 57.3$~\pm~$33.6$^\text{COMBO}$ \\
		diverse-antmaze-med & 0.0 & 27.3$~\pm~$3.9 & \textbf{61.5}$~\pm~$10.0$^\text{CQL}$ & 0.0 \\
		diverse-antmaze-large & 0.0 & \textbf{41.2}$~\pm~$4.2 & 1.0$~\pm~$0.9$^\text{BAIL}$ & 0.0 \\
		\bottomrule
	\end{tabular}
	\label{tab:maze}
\end{table}

\begin{table}
	\centering
	\caption{Performance of ROMI and best performance of prior methods on Gym-MuJoCo tasks.}
	\begin{tabular}{l|c|c|c|c}
		\toprule
		Environment & BC & ROMI-CQL & MF & MB \\
		\midrule
		random-walker2d & 0.0 & 7.5$~\pm~$20.0& \textbf{11.1}$~\pm~$8.8$^\text{ BEAR}$ & 7.0$^\text{ COMBO}$ \\
		random-hopper & 0.9 & \textbf{30.2}$~\pm~$4.4& \textbf{31.4}$~\pm~$0.1$^\text{ CQL}$ & \textbf{31.7}$~\pm~$0.1$^\text{ Repb-SDE}$ \\
		random-halfcheetah & -0.1 & 24.5$~\pm~$0.7& 19.6$~\pm~$1.2$^\text{ CQL}$ & \textbf{38.8}$^\text{ COMBO}$ \\ \midrule
		medium-walker2d & 41.7 & \textbf{84.3}$~\pm~$1.1& \textbf{83.8}$~\pm~$0.2$^\text{ CQL}$ & \textbf{85.3}$~\pm~$2.2$^\text{ Repb-SDE}$ \\
		medium-hopper &  40.0 & 72.3$~\pm~$17.5& 66.6$~\pm~$4.1$^\text{ CQL}$ & \textbf{95.4}$^\text{ MOReL}$ \\
		medium-halfcheetah & 39.2 & 49.1$~\pm~$0.8& 49.0$~\pm~$0.4$^\text{ CQL}$ & \textbf{69.5}$~\pm~$0.0$^\text{ MOPO}$ \\ \midrule
		medium-replay-walker2d & 2.2 & \textbf{109.7}$~\pm~$9.8& 88.4$~\pm~$1.1$^\text{ CQL}$ & 83.8$~\pm~$7.6$^\text{ Repb-SDE}$ \\
		medium-replay-hopper & 8.1 & \textbf{98.1}$~\pm~$2.6& \textbf{97.0}$~\pm~$0.8$^\text{ CQL}$ & 93.6$^\text{ MOReL}$ \\
		medium-replay-halfcheetah & 25.6 & 47.0$~\pm~$0.7 & 46.4$~\pm~$0.3$^\text{ CQL}$ & \textbf{68.2}$~\pm~$3.2$^\text{ MOPO}$ \\ \midrule
		medium-expert-walker2d & 73.4 & \textbf{109.7}$~\pm~$5.3& \textbf{109.5}$~\pm~$0.1$^\text{ CQL}$ & \textbf{111.2}$~\pm~$0.2$^\text{ Repb-SDE}$ \\
		medium-expert-hopper & 36.0 & \textbf{111.4}$~\pm~$5.6& \textbf{106.8}$~\pm~$2.9$^\text{ CQL}$ & \textbf{111.1}$^\text{ COMBO}$ \\
		medium-expert-halfcheetah & 39.7 & 86.8$~\pm~$19.7& 90.8$~\pm~$5.6$^\text{ CQL}$ & \textbf{95.6}$^\text{ MOReL}$ \\
		\bottomrule
	\end{tabular}
	\label{tab:mujoco}
\end{table}

%Table \ref{tab:maze} and \ref{tab:mujoco}summarizes the performance of ROMI and the best performance of prior methods from different categories on the normalized return metric.We categorize prior methods into three types: MF (Model Free), MB (Model Based) and IL (Imitation Learning).
%We also list the performance imporvement of our algorithm over the best MF methods in the corresponding domain.
\subsection{Ablation Study with Model-based Imagination}\label{sec:ablation}

In this subsection, we conduct an ablation study to investigate whether ROMI works due to the reverse model-based imagination. Specifically, we replace the reverse imagination with the forward direction in ROMI, which is denoted as \textit{Forward rOMI (FOMI)}. In this subsection, we study the performance of ROMI and FOMI in \textit{maze2d} and \textit{antmaze} domains and defer the ablation study in \textit{gym} domain to Appendix \ref{appendix_fomi_cql}. Towards fair comparison, we integrate FOMI with BCQ~\cite{fujimoto2019off} in these settings, called \textit{FOMI-BCQ}. Table \ref{tab:abla_maze} shows that ROMI-BCQ significantly outperforms FOMI-BCQ and the base model-free method BCQ~\cite{fujimoto2019off}, which implies that reverse model-based imagination is critical for ROMI in the offline RL settings. 

In the \textit{maze2d} domain, compared with the base model-free algorithm BCQ, ROMI-BCQ outperforms all settings, while FOMI-BCQ achieves the superior performance in \textit{maze2d-medium} but performs poorly in the \textit{umaze} and \textit{large} layouts. As illustrated in Figure~\ref{fig:settings}, \textit{maze2d-medium} enjoys less obstacles on the diagonal to the goal than \textit{maze2d-umaze} and \textit{maze2d-large}. In this case, the conservative reverse model-based imagination can enable safe generalization in all layouts.
%, while the more radical forward imagination can only perform well in \textit{maze2d-medium}. 
A detailed case study of \textit{maze2d-umaze} will be provided in Section \ref{sec:visualization}. Moreover, in the \textit{antmaze} domain, ROMI-BCQ achieve the best performance in the \textit{medium} and \textit{large} layouts, while FOMI-BCQ and BCQ perform well in \textit{antmaze-umaze}. From Figure~\ref{fig:settings}, we find that the \textit{medium} and \textit{large} layouts of \textit{antmaze} have larger mazes with narrower passages, which may frustrate forward imagination and make reverse imagination more effective. 

\begin{table}[h]
	\centering
	\caption{Ablation study about ROMI with model-based imagination. Delta equals the improvement of ROMI-BCQ over BCQ on the normalized return metric.}
	\begin{tabular}{l|l|c|c|c|c}
		\toprule
		Dataset type & Environment &ROMI-BCQ (ours) & FOMI-BCQ & BCQ (base) & Delta \\
		\midrule
		sparse & maze2d-umaze & \textbf{139.5}$~\pm~$3.6 & 8.1$~\pm~$15.5 & 41.1$~\pm~$7.6 & 98.4\\
		sparse & maze2d-medium & 82.4$~\pm~$15.2 & \textbf{93.6}$~\pm~$41.3 & 9.7$~\pm~$14.2 & 72.7\\
		sparse & maze2d-large & \textbf{83.1}$~\pm~$22.1 & -2.5$~\pm~$0.0 & 38.3$~\pm~$10.4 & 44.8\\
		dense & maze2d-umaze & \textbf{98.3}$~\pm~$2.5 & 30.7$~\pm~$0.9 & 37.0$~\pm~$5.3 & 61.3\\
		dense & maze2d-medium & \textbf{102.6}$~\pm~$32.4 & 64.7$~\pm~$37.0 & 37.9$~\pm~$4.5 & 64.7\\
		dense & maze2d-large & \textbf{124.0}$~\pm~$1.3& -0.7$~\pm~$7.1 & 79.8$~\pm~$12.2 & 44.2\\ \midrule
		fixed & antmaze-umaze & 68.7$~\pm~$2.7 & \textbf{79.5}$~\pm~$2.5 & 75.3$~\pm~$13.7 & -6.6\\
		play & antmaze-medium & \textbf{35.3}$~\pm~$1.3 & 26.2$~\pm~$5.5 & 0.0 & 35.3\\
		play & antmaze-large & \textbf{20.2}$~\pm~$14.8 & 12.0$~\pm~$3.3 & 0.0 & 20.2\\
		diverse & antmaze-umaze & 61.2$~\pm~$3.3 & \textbf{66.8}$~\pm~$3.5 & 49.3$~\pm~$9.9 & 11.9\\
		diverse & antmaze-medium & \textbf{27.3}$~\pm~$3.9 & 12.3$~\pm~$2.1 & 0.0 & 27.3\\
		diverse & antmaze-large & \textbf{41.2}$~\pm~$4.2 & 17.8$~\pm~$2.1 & 0.0 & 41.2\\
		\bottomrule
	\end{tabular}
	\label{tab:abla_maze}
\end{table}

\subsection{Ablation Study with Different Rollout Policies}
In this subsection, we conduct an ablation study to investigate the effect of ROMI's different rollout policies. To compare with a CVAE-based policy, we propose a new rollout policy (i.e., reverse behavior cloning) for ROMI-BCQ, denoted by \textit{ROMI-RBC-BCQ}. To realize the reverse behavior cloning method, we train a stochastic policy $\widehat{\pi}_\varphi(a|s')$, which is parameterized by $\varphi$ and can sample current action depended on the next state. During training $\widehat{\pi}_\varphi(a|s')$, the objective $\gL_{rbc}(\varphi)$ is formalized by
\begin{align}
	\gL_{rbc}(\varphi) = \mathop{\E}_{(s,a,r,s')\sim \Denv}\left[-\log \widehat{\pi}_\varphi(a|s')\right],
\end{align}
where $\Denv$ is the given offline dataset, and minimizing the loss function $\gL_{rbc}(\varphi)$ is equivalent to maximizing the log-likelihood of probability.

We illustrate the performance of ROMI with different rollout policies in Table~\ref{tab:abla_rbc}, where ROMI-BCQ achieves the best performance and ROMI-RBC-BCQ also outperforms BCQ. As suggested by prior generative models~\citep{kingma2013auto,sohn2015learning}, in comparison to the deterministic layers (e.g., fully connected layers), CVAE-based methods can generate more diverse and realistic structured samples using stochastic inference. We argue that, the rollout policy implemented by CVAE is critical for ROMI to rollout more diverse trajectories for proposed generalization, and BC-based implementation is also effective in reverse model-based imagination. 

\begin{table}[h]
	\centering
	\caption{Ablation study about ROMI with different rollout policies.}
	\begin{tabular}{l|l|c|c|c}
		\toprule
		Dataset type & Environment & ROMI-BCQ (ours) & ROMI-RBC-BCQ & BCQ (base)  \\
		\midrule
		fixed & antmaze-umaze & 68.7$~\pm~$2.7 & 62.2$~\pm~$5.6 & \textbf{75.3}$~\pm~$13.7 \\
		play & antmaze-medium & \textbf{35.3}$~\pm~$1.3 & \textbf{33.8}$~\pm~$6.2 & 0.0 \\
		play & antmaze-large & \textbf{20.2}$~\pm~$14.8 & 13.3$~\pm~$16.1 & 0.0 \\
		diverse & antmaze-umaze & \textbf{61.2}$~\pm~$3.3 & 43.8$~\pm~$13.3 & 49.3$~\pm~$9.9 \\
		diverse & antmaze-medium & \textbf{27.3}$~\pm~$3.9 & 20.8$~\pm~$15.5 & 0.0 \\
		diverse & antmaze-large & \textbf{41.2}$~\pm~$4.2 & 14.2$~\pm~$9.8 & 0.0 \\
		\bottomrule
	\end{tabular}
	\label{tab:abla_rbc}
\end{table}

\subsection{A Case Study in \textit{maze2d-umaze}}
\label{sec:visualization}

To dive deeper into how ROMI triggers more conservative and effective behaviors, we provide a detailed visual demonstration of one particular task in D4RL benchmark~\citep{fu2020d4rl}: \textit{maze2d-umaze-sparse}.

\begin{figure}[h]
	\centering
	\vspace{-0.2in}
	\includegraphics[width=0.97\linewidth]{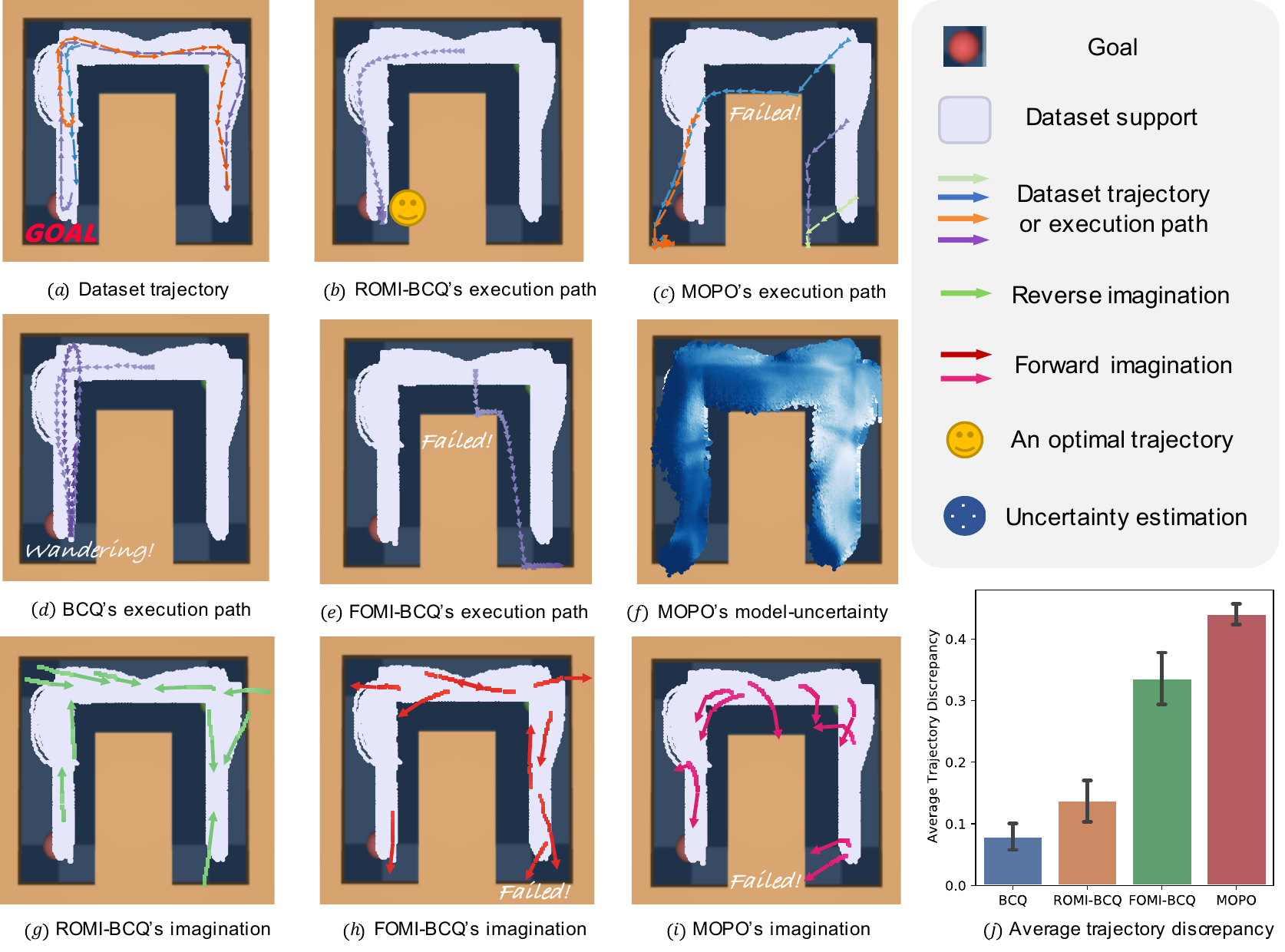}
	\caption{A case study in \textit{maze2d-umaze}. Note that in figure (f), the penalty is reflected by the color, where blue represents large penalty while white represents small penalty.}
	\label{fig:visualization}
\end{figure}

As mentioned in Section~\ref{sec:environments}, experiences in \textit{maze2d} domain are generated by a planner moving between randomly sampled waypoints on the clearing. Figure~\ref{fig:visualization}a shows the movement of the agent from randomly sampled trajectories. To earn high returns, the agent not only needs to learn how to direct to the goal, but also how to stay in the high reward region --- the latter behavior is not in the dataset yet. Model-free offline RL algorithms constrain their policy "close" to the dataset, thus it is hard to learn such behaviors out of the support. To see this, Figure~\ref{fig:visualization}d shows the behavior of a trained BCQ policy during execution. After reaching the goal, the agent will still oscillate between the high-reward and low-reward regions. This motivates us to use model imagination to generalize beyond the dataset.

As shown in Table~\ref{tab:maze}, ROMI solves this task but previous model-based methods have poor performance, sometimes even worse than model-free algorithms. To better understand this counter-intuitive phenomenon, we compare the rollout trajectories and the learned policy of ROMI-BCQ, FOMI-BCQ (mentioned in Section \ref{sec:ablation}), and MOPO~\cite{yu2020mopo}. Figure~\ref{fig:visualization}(g-i) shows the imagined trajectories in the learning process of the three methods. Figure~\ref{fig:visualization}(b,c,e) shows the learned policy behavior at the execution phase. While all model-based imagination will leave the dataset for better generalization, forward model rollout naturally takes some risks
%lose some conservatism 
as it directs the agent to unknown areas. Undesired forward model imagination will ruin the policy learning (e.g., FOMI-BCQ in Figure~\ref{fig:visualization}e and Table~\ref{tab:abla_maze}) or mislead the policy optimization to the suboptimal solution (e.g., MOPO in Figure~\ref{fig:visualization}c). Moreover, as shown in Figure~\ref{fig:visualization}f, the regularization penalty based on model uncertainty also failed, which is also pointed out in the literature~\cite{yu2021combo}.
%\wl{do we need to cite COMBO here?}. 
On the other hand, reverse model imagination inherits the conservatism of the dataset, as it is always ended in the real experience points. Figure~\ref{fig:visualization}(b,g) shows that ROMI induces the conservative and optimal behaviors:
%realizes the conservatism in model imagination and learns the optimal policy: 
ROMI will stay around the goal point, and will stick to the data points for higher expected returns. We aim to quantify the aggressiveness of each learned policy and thus define the following trajectory-based discrepancy to measure the distance between learning policy and dataset:
%when generalization is unnecessary. 
%To quantify the distance between policy behavior and dataset, we define the following trajectory-based discrepancy:
\begin{definition}[Average Trajectory Discrepancy]
	Given a dataset $\gD=\{(s,a,r,s^\prime)\}$ and a trajectory $\tau=(s_0,a_0,\dots,s_H,a_H)$, the discrepancy between $\gD$ and $\tau$ is defined as:
	\begin{equation}\label{eq:atd}
	\sD(\gD, \tau)=\frac{1}{H+1}\sum_{t=0}^H\min_{(s,a,r,s^\prime)\in\gD}\Vert s_t-s\Vert_2.
	\end{equation}
\end{definition}
We report the average trajectory discrepancy for policies during execution in Figure~\ref{fig:visualization}j and defer results for other maze environments in Appendix~\ref{appendix_discrepancy}. The results implicate that ROMI is as conservative as model-free method BCQ, but forward model-based offline RL policy (i.e., ROMI and MOPO) tends to deviate from the dataset and will touch the undesirable walls.
%lead to undesirable results.

%% file: 5-Related-Work.tex
\section{Related Work}
\label{sec:relatedwork}
%\wl{
%\begin{itemize}
%    \item offline RL: incorporating conservatism
%    \item model-based RL: recall traces
%\end{itemize}
%}

\textbf{Offline RL.} In the literature, offline RL methods can be primarily divided into model-free and model-based algorithms. Prior model-free offline RL methods constrain their policy search in the offline dataset. They can realize their bias by either estimating uncertainty quantification to the value function~\cite{levine2020offline,wu2019behavior,kumar2019stabilizing,agarwal2020optimistic}, using importance sampling based algorithms~\cite{precup2001off,sutton2016emphatic,liu2020off,swaminathan2015batch,nachum2019algaedice,singh2020cog,peng2019advantage}, explicitly constraining the learning policy to be close to the dataset~\cite{fujimoto2019off,liu2020provably}, learning the conservative value function~\cite{kumar2020conservative}, and using KL divergence~\cite{jaques2019way,wu2019behavior,zhou2020plas} or MMD~\cite{kumar2019stabilizing}. On the other hand, prior model-based offline RL methods~\cite{kidambi2020morel,yu2020mopo,lee2021representation,finn2017deep,ebert2018visual,kahn2018composable,matsushima2020deployment,argenson2020model,swazinna2020overcoming,rafailov2020offline} have studied in model-uncertainty quantification~\cite{kidambi2020morel,yu2020mopo,ovadia2019can}, representation learning~\cite{lee2021representation}, constraining the policy to imitate the behavioral policy~\cite{matsushima2020deployment}, and using conservative estimation of value function~\cite{yu2021combo}. Different from these works, we propose ROMI to investigate a new direction in the model-based offline RL, which will provide natural conservatism bias with maintaining superior model-based generalization benefits.

%Prior model-free offline RL algorithms have been designed to regularize the learned policy to be “close“ to the behavioral policy either implicitly via regularized variants of importance sampling based algorithms, offline actor-critic methods, applying uncertainty quantification to the pre- dictions of the Q-values, and learning conservative Q-values or explicitly measured by direct state or action constraints, KL divergence, Wasserstein distance, and MMD.

%Recent studies~\citep{} have shown that standard off-policy methods lead to poor results on fixed datasets. The cause of these failures is identified as extrapolation or bootstrapping errors~\citep{}: the Q-function fails to provide accurate estimations for state-action pairs not present in the fixed dataset.

%To address this challenge, previous algorithms follow the paradigm of incorporating conservatism or regularization to the policy learning. Model-free approaches~\cite{} incorporate conservatism by constraining the learned policy to be "close" to the behavior policy or penalizing overly optimistic Q-values in the Bellman update. To achieve better generalization, model-based approaches~\cite{} learn a dynamic model to generate new samples, and incorporate conservatism via Q-value penalty or uncertainty quantification.

%The aforementioned offline RL algorithms incorporate conservatism to the policy optimization. Alternatively, we build conservatism in the dataset, which can be applied to various policy optimization methods for different scenarios.

\textbf{Reverse model-based RL.} The idea of learning a reverse dynamics model to imagine reversal samples from goal states first emerges in the literature of online RL algorithms~\cite{holyoak1999bidirectional,goyal2018recall,edwards2018forward,lai2020bidirectional,lee2020context}, and shows the potential of speeding up learning and improving sample efficiency by aimed exploration. Similar backward model shows benefit in planning for credit assignment~\cite{van2019use, chelu2020forethought} and robustness~\cite{jafferjee2020hallucinating}. Lai et al. (2020)~\cite{lai2020bidirectional} utilizes a backward model to reduce the reliance on accuracy in forward model predictions. In contrast to the backward model in online RL, we show that reverse imagination inherently incorporates the conservatism into rollout trajectories in the offline RL, which can lead rollouts towards target states in dataset. To our best knowledge, ROMI is the first offline RL method with reverse model-based imagination to induce conservatism bias with data augmentation.

%% file: 6-Conclusion.tex
\vspace{-0.1in}
\section{Conclusion}
\label{sec:conclusion}
This paper introduces ROMI, a novel model-based framework for offline RL. To enable conservative generalization, it adopts reverse dynamics model to imagine possible trajectories that can lead to the states within the offline dataset. We demonstrate ROMI can effectively combine with off-the-shelf model-free algorithms and achieve state-of-the-art performance on offline RL benchmark tasks. ROMI leverages good trajectories (e.g., reaching goal states or with high rewards) in an offline dataset to generate effective augmentation data. Although most practical offline datasets contain such trajectories, it is challenging for ROMI to work in other rare cases. One potential future direction is to extend ROMI with forward models, generating bolder imaginations and reaching goal states outside a given dataset. Theoretical formalization of the reverse imagination in offline RL is also an interesting future direction.

%% file: 7-Appendix.tex
\appendix
\section{An Illustrative Example for ROMI}\label{appendix_example}
Assume state $s_\text{in}$ has a trajectory leading to the goal (denoted by $s_\text{goal}$) in the dataset. Considered below are two five-step trajectories from forward imagination and reverse imagination, which visit the same state sequence, but in the opposite direction (hence the actions are not the same). Let the forward imaginary trajectory be from $s_\text{in}$ to $s_5$, where $s_5$ is the end state of the trajectory. Similarly, let the reverse imaginary trajectory be from $s_5$ to $s_\text{in}$. Note that the rollout orders of these two trajectories are the same, starting at $s_\text{in}$ and ending at $s_5$. We formalize these two imaginary trajectories as follows:

\textbf{Forward rollout: } $\langle s_\text{in},a_1^f,s_1,a_2^f,s_2,a_3^f,s_3,a_4^f,s_4,a_5^f,s_5\rangle$,

\textbf{Reverse rollout: } $\langle s_5,a_5^r,s_4,a_4^r,s_3,a_3^r,s_2,a_2^r,s_1,a_1^r,s_\text{in}\rangle$,

   where $s_{1:5}$ denotes the imaginary state sequence, $a_{1:5}^f$ denotes the action sequence for forward imagination, and $a_{1:5}^r$ denotes the action sequence for reverse imagination.

   During the training process, the reverse rollout will expand the trajectory from $\langle s_\text{in},\cdots,s_\text{goal}\rangle$ to $\langle s_5,\cdots,s_\text{in},\cdots,s_\text{goal}\rangle$, i.e., $s_{1:5}$ can now reach the goal. Through the reverse imagination of $s_{1:5}$, the reverse rollout can benefit the policy learning in this task. For the forward rollout, there are three cases for the state $s_5$:

\begin{enumerate}
    \item Consider $s_5$ is outside the dataset and the value of $s_5$ (i.e., $Q(s_5,\cdot)$) is overestimated, which is a common challenge for function approximation in offline RL~\cite{fujimoto2019off}. The forward rollout can mislead the learning policy from $s_\text{in}$ to $s_5$ and harm the learning performance. In contrast, the reverse rollout does not have such negative impact. It is because that $s_\text{in}$ does not have an action sequence to reach $s_5$ in the reversed data augmentation.
    \item If $s_5$ is outside the dataset and the value of $s_5$ is not overestimated, the forward rollout does not affect the policy learning. It is because that the execution policy will not go to the state $s_5$ with a lower $Q(s_5,\cdot)$). In contrast, as discussed above, the reverse rollout would benefit the learning in this case due to its effective trajectory expansion with the opposite direction of imagination.
    \item Consider $s_5$ is inside the dataset. When $s_5$ has a trajectory leading to the goal in the dataset, we believe the forward rollout can improve the learning by selecting the better of the two successful trajectories, $\langle s_\text{in},\cdots,s_\text{goal}\rangle$ and $\langle s_\text{in},\cdots,s_5,\cdots,s_\text{goal}\rangle$. In the reversed data augmentation, there is only one successful trajectory from the state $s_\text{in}$, i.e., $\langle s_\text{in},\cdots,s_\text{goal}\rangle$, but this could naturally be compensated by generating a reverse rollout from state $s_5$. Note that, in this case, both forward and reverse methods do not need to deal with the conservatism issue, because $s_5$ is in the original offline dataset.
\end{enumerate}

   In summary, we would like to highlight that reverse imagination is more conservative for imagined states outside the dataset and may even have overly estimated values. This explains why it works well in offline RL.

\section{Omitted Background in Section 3}\label{appendix_CVAE}
\paragraph{Conditional Variational Auto-Encoder}
A variational auto-encoder (VAE)~\citep{kingma2013auto} is a generative model of $p_\theta(X) = \prod_{i=1}^{n}p_\theta(x_i)$, where $X = \{x_1, \dots, x_n\}$ denotes the given dataset and $\theta$ denotes the parameters of approximate ML or MAP estimation. It leverages the latent variable model structure of $p_\theta(X)=\int p_\theta(X|z)p(z)dz,$
where $z$ is the latent variable with prior distribution $p(z)$. To perform tractable maximum log-likelihood, VAE optimizes the following variational lower bound:
\begin{align}
\log p_\theta(X)\geq\E_{z\sim q_\xi(\cdot|X)}[\log p_\theta(X|z)]-D_{\text{KL}}\left(q_\xi(z|X)\Vert p(z)\right),
\end{align}
where $q_\xi(z|X)=\gN\left(z|\bm{\mu}_\xi(X),\bm{\Sigma}_\xi(X)\right)$ is the encoder parameterized by $\xi$ that represents a multivariate Gaussian distribution with mean $\bm{\mu}_\xi$ and variance $\bm{\Sigma}_\xi$, and $p_\theta(X|z)$ is the decoder parameterized by $\theta$. With reparametrization trick~\citep{kingma2013auto}, VAE performs gradient descent on the variational lower bound. For inference, $z$ is sampled from a Gaussian distribution and passed through the decoder to generate diverse samples $x$.

Conditional VAE (CVAE)~\citep{sohn2015learning} is a variant of VAE , which aims to model $p_\theta(Y|X)$. Similar to variational lower bound, CVAE optimizes the following lower bound:
\begin{align}
\log p_\theta(Y|X)\geq\E_{z\sim q_\xi(\cdot|Y,X)}\left[\log p_\theta(Y|z,X)\right]-D_{\text{KL}}\left(q_\xi(z|Y,X)\Vert p(z|X)\right),
\end{align}
where $q_\xi(z|Y,X),p_\theta(Y|z,X)$ also denote the encoder and decoder parameterized by $\xi,\theta$ with the evidence $X$, respectively.

\section{Omitted Quantification of Model Accuracy in Section 4.1}\label{appendix_model_acc}
To compare the model accuracy of forward and reverse imagination, we estimate the model error of forward and reverse models. To be fair, we utilize the same neural network architecture to implement these models. We quantify the one-step model error as mean squared error (MSE) on the validation set.
% follows:
% \begin{definition}[One-step Model Error]
% 	Given a dataset $\Denv=\{(s,a,r,s^\prime)\}$, the one-step forward model error $\epsilon_{fm}$ and reverse model error $\epsilon_{rm}$ are defined as:
% 	\begin{align}
% 	\epsilon_{fm} &= \mathop{\E}_{(s,a,r,s')\sim \Denv,\hat{s}'\sim\widehat{T}_f(\cdot|s,a)}\left[\Vert s' - \hat{s}'\Vert_2^2+\left(r-\widehat{r}(s,a)\right)^2\right] \\
% 	\epsilon_{rm} &= \mathop{\E}_{(s,a,r,s')\sim \Denv,\left(\tilde{s},\tilde{r}\right)\sim \widehat{p}_\phi(\cdot|s',a),\tilde{s}'\sim T(\cdot|\tilde{s},a)}\left[\Vert s' - \tilde{s}'\Vert_2^2+\left(r-\tilde{r}\right)^2\right],
% 	\end{align}
% 	where $\widehat{T}_f,\widehat{r}$ denote the approximated forward dynamics and reward model introduced in Section 2 respectively, and $\widehat{p}_\phi$ denotes the learned reverse dynamics and reward model discussed in Section 3.
% \end{definition}
% $s_\text{in}$ce the oracle reverse transition function $T_r(s|s',a)=T^{-1}(s'|s,a)$ is basically inaccessible in practice, in the definition of $\epsilon_{rm}$, we rollout the approximated reverse model $\widehat{p}_\phi$ at first, and sample the next state $\tilde{s}'$ u$s_\text{in}$g oracle forward transition function $T$ to calculate the one-step reverse model error. 
We evaluate the forward and reverse model error in five different robots of D4RL benchmark suite~\citep{fu2020d4rl} illustrated in Table~\ref{tab:modelerror}. For each robot, we report the MSE across different types of configurations. Table~\ref{tab:modelerror} shows that reverse models have comparable accuracy to, if not worse than, forward models.
%We defer the detailed measure of the model accuracy quantification to Appendix. 
%Moreover, to generate diverse imaginary trajectories in complex tasks, we adopt a conditional VAE to realize ROMI's rollout policy, which has been introduced in Section~3. We observe that ROMI's reverse model is quite accurate in the \textit{maze2d} and \textit{hopper} domains, in which ROMI can also use a uniform policy to launch diverse trajectories instead. We observe that ROMI's reverse model is quite accurate in the \textit{maze2d} and \textit{hopper} domains, in which ROMI can also use a uniform policy to launch diverse trajectories instead.

% \begin{table}[h]
% 	\caption{One-step model error of forward model and reverse model.}
% 	\label{tab:modelerror}
% 	\centering
% 	\begin{tabular}{c|c|c|c|c|c}
% 		\toprule
% 		Model type & maze2d & antmaze & walker2d & hopper & halfcheetah \\
% 		\midrule
% 		Forward model & 0.0051 & 0.1178 & 0.7308 & 0.0098 & 0.2650 \\
% 		Reverse model & 0.0097 & 0.0945 & 0.7163 & 0.0092 & 0.3967 \\
% 		\bottomrule
% 	\end{tabular}
% \end{table}

\begin{table}[H]
	\caption{MSE of forward model and reverse model.}
	\label{tab:modelerror}
	\centering
	\begin{tabular}{l|l|c|c}
		\toprule
		Dataset type & Environment & Forward model & Reverse model \\
		\midrule
		sparse & maze2d-umaze & 0.0062$~\pm~$0.0019 & 0.0061$~\pm~$0.0019 \\
		sparse & maze2d-medium & 0.0087$~\pm~$0.0017 & 0.0107$~\pm~$0.0011 \\
		sparse & maze2d-large & 0.0195$~\pm~$0.0083 & 0.0242$~\pm~$0.0053 \\
		dense & maze2d-umaze & 0.0014$~\pm~$0.0005 & 0.0014$~\pm~$0.0004 \\ 
		dense & maze2d-medium & 0.0008$~\pm~$0.0003 & 0.0007$~\pm~$0.0003 \\
		dense & maze2d-large & 0.0011$~\pm~$0.0004 & 0.0021$~\pm~$0.0001 \\ \midrule
		fixed & antmaze-umaze & 0.1460$~\pm~$0.0470 & 0.2027$~\pm~$0.0086 \\ 
		play & antmaze-medium & 0.1225$~\pm~$0.0212 & 0.1947$~\pm~$0.0053 \\
		play & antmaze-large & 0.1201$~\pm~$0.0272 & 0.1875$~\pm~$0.0058 \\
		diverse & antmaze-umaze & 0.1509$~\pm~$0.0494 & 0.1374$~\pm~$0.0043 \\ 
		diverse & antmaze-medium & 0.1303$~\pm~$0.0206 & 0.1743$~\pm~$0.0139 \\
		diverse & antmaze-large & 0.1167$~\pm~$0.0236 & 0.1813$~\pm~$0.0031 \\ \midrule
		random & mujoco-walker2d & 0.5141$~\pm~$0.0083 & 0.7102$~\pm~$0.0720 \\
		random & mujoco-hopper & 0.0010$~\pm~$0.0004 & 0.0012$~\pm~$0.0006 \\
		random & mujoco-halfcheetah & 0.1268$~\pm~$0.0111 & 1.4220$~\pm~$0.1144 \\
		medium & mujoco-walker2d & 0.2647$~\pm~$0.0680 & 0.2312$~\pm~$0.0048 \\
		medium & mujoco-hopper & 0.0020$~\pm~$0.0002 & 0.0024$~\pm~$0.0002 \\
		medium & mujoco-halfcheetah & 0.2437$~\pm~$0.0156 & 0.5865$~\pm~$0.0247 \\
		medium-replay & mujoco-walker2d & 0.3572$~\pm~$0.0556 & 0.6123$~\pm~$0.0913 \\
		medium-replay & mujoco-hopper & 0.0037$~\pm~$0.0005 & 0.0045$~\pm~$0.0004 \\
		medium-replay & mujoco-halfcheetah & 0.4684$~\pm~$0.0156 & 1.4916$~\pm~$0.1705 \\
		medium-expert & mujoco-walker2d & 0.1447$~\pm~$0.0157 & 0.1687$~\pm~$0.0107 \\
		medium-expert & mujoco-hopper & 0.0022$~\pm~$0.0004 & 0.0019$~\pm~$0.0005 \\
		medium-expert & mujoco-halfcheetah & 0.1586$~\pm~$0.0093 & 0.5161$~\pm~$0.0389 \\
		\bottomrule
	\end{tabular}
\end{table}

\section{Experiment Settings and Implementation Details}

\subsection{Evaluated Settings of D4RL Benchmark}

We consider a wide range of domains in the popular offline RL benchmark, D4RL~\cite{fu2020d4rl}. D4RL is specifically designed for offline RL setting, which contains many elaborate datasets with key properties related to the practical offline RL applications. As discussed in Section 4.1, we use three D4RL benchmark domains with 24 datasets by controlling five different robots. We adopt the normalized scores metric suggested by D4RL benchmark~\cite{fu2020d4rl}, where 0 indicates a random policy performance and 100 corresponds to an expert. As suggested by (Fu et al., 2020)~\cite{fu2020d4rl}, we illustrate the reference min and max scores of our evaluated D4RL settings in Table \ref{tab:d4rl_score_range}, where score $C$ is normalized to $\widetilde{C}$ by
\begin{align}
	\widetilde{C} = \frac{C - \text{reference min score}}{\text{reference max score} - \text{reference min score}}.
\end{align}
In the \textit{maze2d} domain, different types of datasets have different reference scores. Moreover, in the \textit{antmaze}, \textit{walker2d} \textit{hopper}, and \textit{halfcheetah} domains, all domains share the same reference min and max scores. We assume that the termination conditions of tasks are known, which is common in other model-based offline RL baselines~\cite{yu2020mopo,lee2021representation}. Note that we evaluate ROMI with all baselines in the \textit{*-v2} datasets of \textit{gym} domain (i.e., \textit{walker2d}, \textit{hopper}, and \textit{halfcheetah}). Table \ref{tab:d4rl_score_range} also shows the episode length of each task, in which the algorithms will time-out when the execution phase exceeds the time limit of each episode.
 
\begin{table}[H]
	\centering
	\caption{Reference minmax scores and episode lengths of evaluated D4RL settings.}
	\label{tab:d4rl_score_range}
	\begin{tabular}{l|l|c|c|c}
		\toprule
		Dataset type & Environment & Reference min score & Reference max score & Ep. length \\
		\midrule
		sparse & maze2d-umaze & 23.85 & 161.86 & 300 \\
		sparse & maze2d-medium & 13.13 & 277.39 & 600 \\
		sparse & maze2d-large & 6.7 & 273.99 & 800 \\
		dense & maze2d-umaze & 68.54  & 193.66 & 300 \\ 
		dense & maze2d-medium & 44.26 & 297.46 & 600 \\
		dense & maze2d-large & 30.57 & 303.49 & 800 \\ \midrule
		\textemdash & antmaze-umaze & 0.0 & 1.0 & 700 \\ 
		\textemdash & antmaze-medium & 0.0 & 1.0 & 1000 \\
		\textemdash & antmaze-large & 0.0 & 1.0 & 1000 \\ \midrule
		\textemdash & walker2d & 1.63 & 4592.3 & 1000 \\
		\textemdash & hopper & -20.27 & 3234.3 & 1000 \\
		\textemdash & halfcheetah & -280.18 & 12135.0 & 1000 \\
		\bottomrule
	\end{tabular}
\end{table}
 
\subsection{Implmentation Details}\label{appendix_alg_details}

We train a probabilistic neural network to represent the approximated reverse dynamics and reward model, which takes the next state and current action as input and outputs a diagonal multivariate Gaussian distribution predicting state difference and reward: 
\begin{align}
	\widehat{p}_\phi(s, r|s', a) = \gN(\bm{\mu}_\phi(s',a),\bm{\Sigma}_\phi(s',a)),
\end{align}
where $\bm{\mu}_\phi$ and $\bm{\Sigma}_\phi$ denote the mean and variance of reverse model $\widehat{p}_\phi$, respectively. To boost the model accuracy, following Janner et al. (2019)~\cite{janner2019trust}, we train an ensemble of seven such models and pick the best five models according to the validated performance on a hold-out set of 1000 transitions in the offline dataset. During the model-based imagination, we randomly select a model from the best five model candidates to rollout trajectories per step. The neural network architecture of each model has four feedforward layers with 200 hidden units. Each intermediate layer has a \textit{swish} activation, and the inputs and outputs of our models are normalized across the given dataset.

%During model rollouts, we randomly pick one dynamics model from the best 5 models. 
%Each model in the ensemble is represented as a 4-layer feedforward neural network with 200 hidden units. 

%based on the validation prediction error on a held-out set that contains 1000 transitions in the offline dataset.

%Across all domains, we train an ensemble of 7 models and pick the best 5 models based on their prediction error on a hold-out set of 1000 transitions in the offline dataset. 

%Each of the models in the ensemble is parametrized as a 4-layer feedforward neural network with 200 hidden units.

%We use 200 hidden units for all intermediate layers. 
%Across all domains, we train an ensemble of 7 models and pick the best 5 models on their validation error on a hold-out set of 1000 transitions in the dataset. 
%The inputs and outputs of the neural network are normalized. 

The rollout policy $\widehat{G}_\theta(a|s')$ implemented by a conditional VAE has two components: an encoder $\widehat{E}_\omega(s',a)$ and a decoder $\widehat{D}_\xi(s',z)$. Both the encoder and the decoder contain two intermediate layers with 750 hidden units and \textit{relu} activations. Besides the common implementation of conditional VAE~\cite{sohn2015learning,fujimoto2019off}, our rollout policy $\widehat{G}_\theta(a|s')$ takes the next state as input and can sample reverse action u$s_\text{in}$g stochastic inference from an underlying latent space. To generate diverse imaginary trajectories in complex tasks, we adopt such conditional VAE as a rollout policy introduced in Section~3. Moreover, we observe that ROMI's reverse model is quite accurate in the \textit{maze2d} and \textit{hopper} domains, in which ROMI can also use a uniform policy to launch diverse trajectories instead. We introduce the hyperparameters of ROMI in Table~\ref{tab:hyperparameter}. As suggested by the evaluated reverse model error in Table~\ref{tab:modelerror}, we perform short-horizon model rollouts in the 24 tasks of D4RL benchmark, in which ROMI rollouts five steps in domains with higher model accuracy and performs one step imagination in \textit{walker2d} and \textit{halfcheetah}. For \emph{antmaze} with sparse rewards, we prioritize the start point of model imagination, in which the prioritized weights are similarly introduced in BAIL~\cite{chen2020bail}.

We combine ROMI with prior offline RL learning algorithms (i.e., BCQ~\cite{fujimoto2019off} or CQL~\cite{kumar2020conservative}) to conduct extensive evaluations on D4RL benchmark suite. As discussed in Section 3, ROMI combines the given dataset $\Denv$ and model-based buffer $\gD_{\text{model}}$ to obtain the total dataset $\gD_{\text{total}}$ for model-free learning methods. During policy training, we collect the training minibatches from two sources, $\Denv$ and $\gD_{\text{model}}$, and define the ratio of data collected from the model-based buffer $\gD_{\text{model}}$ as $\eta \in [0, 1]$. Therefore, the ratio of data collected from $\Denv$ is naturally derived by $1-\eta$. We list the hyperparameters $\eta$ in Table~\ref{tab:hyperparameter}, where we search over $\eta \in \{0.1, 0.3, 0.5, 0.7, 0.9\}$ and choose the best one for each domain. We found that data ratio $\eta=0.1$ is generally effective for ROMI. The only exception is the \textit{maze2d} domain, in which other ratios were found to work better. Moreover, ROMI's training time on an NVIDIA RTX 2080TI GPU of each task is about eight hours to 20 hours with the BCQ or CQL learning algorithms. 

We compare ROMI with eleven state-of-the-art offline RL baselines: BC (behavior cloning), BCQ~\cite{fujimoto2019off}, BEAR~\cite{kumar2019stabilizing}, BRAC-v, BRAC-p~\cite{wu2019behavior}, BAIL~\cite{chen2020bail}, CQL~\cite{kumar2020conservative}, MOPO~\cite{yu2020mopo}, MOReL \citep{kidambi2020morel}, Repb-SDE~\cite{lee2021representation}, and COMBO \citep{yu2021combo}. Our implementation of these baseline algorithms refers to their public source codes. In particularly, we utilize the implementation of BC introduced by the Repb-SDE repository and the implementation of MOReL and COMBO is referred to \url{https://github.com/SwapnilPande/MOReL/} and \url{https://github.com/takuseno/d3rlpy/}, respectively.

%In this section, we discuss the hyperparameters associated with COMBO along with guidelines to choose hyperparameters for offline RL tasks with a lot of variety in terms of dataset diversity, input modality, etc. 
%We now list the additional hyperparameters and our decisions behind these (which can serve as guidelines/rules for hyperparameter selection when COMBO is used on a new task) as follows.

%QPLEX is also based on PyMARL, whose special hyper-parameters are illustrated in Table 2 and other common hyper-parameters are adopted by the default implementation of PyMARL (Samvelyan et al., 2019). 

\begin{table}[H]
	\caption{Hyperparameters of ROMI used in the D4RL datasets.}
	\label{tab:hyperparameter}
	\centering
	\begin{tabular}{l|c|c|l}
		\toprule
		Environment & Rollout length & Data ratio $\eta$ & Rollout policy  \\
		\midrule
		maze2d-umaze & 5 & 0.7 & uniform \\
		maze2d-medium & 5 & 0.5 & uniform \\
		maze2d-large & 5 & 0.3 & uniform \\ \midrule
		antmaze-umaze & 5 & 0.1 & conditional VAE \\ 
		antmaze-medium & 5 & 0.1& conditional VAE \\
		antmaze-large & 5 & 0.1 & conditional VAE \\ \midrule
		walker2d & 1 & 0.1 & conditional VAE \\
		hopper & 5 & 0.1 & uniform \\
		halfcheetah & 1 & 0.1 & conditional VAE \\
		\bottomrule
	\end{tabular}
\end{table}

\section{Omitted Tables in Section 4.2}\label{appendix_full_exp}
See Table~\ref{large}.

\addtolength{\tabcolsep}{-1pt}

\begin{sidewaystable}[htbp]
	\caption{Performance of ROMI with eleven offline RL baselines on the \textit{maze}, \textit{antmaze}, \textit{walker2d}, \textit{hopper}, \textit{halfcheetah} domains with the normalized score metric proposed by D4RL benchmark~\citep{fu2020d4rl}, averaged over three random seeds with $\pm$ standard deviation. Scores roughly range from 0 to 100, where 0 corresponds to a random policy performance and 100 indicates an expert.}
	\label{large}
	\begin{center}
		\scriptsize
		\begin{tabular}{l|l|c|c|c|cccccc|ccccc}
			\toprule
			Dataset type & Environment & BC & ROMI-BCQ (ours) & ROMI-CQL (ours) & BCQ & CQL & BEAR & BRAC-p & BRAC-v & BAIL & MOPO & MOReL & Repb-SDE & COMBO\\ \midrule
			sparse & maze2d-umaze & -3.2 & 139.5$~\pm~$3.6 & 54.3$~\pm~$12.2 & 49.1 & 18.9 & 65.7 & -9.2 & 8.7 & 44.7 & -14.8 & -14.4 & -12.3 & 76.4\\
			sparse & maze2d-med & -0.5 & 82.4$~\pm~$15.2 & 13.3$~\pm~$10.0 & 17.1 & 14.6 & 25.0 & 70.0 & 70.6 & 19.7 & 37.4 & -4.8 & -4.7 & 68.5\\
			sparse & maze2d-large & -1.7 & 83.1$~\pm~$22.1 & 10.2$~\pm~$5.2 & 30.8 & 16.0 & 81.0 & 0.2 & -2.2 & 4.4 & -0.8 & -2.3 & -2.2 & 14.1\\
			dense & maze2d-umaze & -6.9 & 98.3$~\pm~$2.5 & 39.4$~\pm~$11.8 & 48.4 & 14.4 & 32.6 & 51.5 & 17.6 & 43.3 & 29.9 & 36.9 & 94.3 & 86.2\\
			dense & maze2d-med & 2.7 & 102.6$~\pm~$32.4 & 37.0$~\pm~$4.1 & 41.1 & 30.5 & 19.1 & 73.0 & 24.4 & 41.7 & 70.5 & -1.9 & 22.7 & 84.2\\ 
			dense & maze2d-large & -0.3 & 124.0$~\pm~$1.3 & 44.0$~\pm~$1.9 & 75.2 & 46.9 & 133.8 & -4.7 & -7.0 & 58.2 & 36.8 & -3.2 & 7.9 & 100.2\\ \midrule
			fixed & antmaze-umaze & 82.0 & 68.7$~\pm~$2.7 & 68.1$~\pm~$5.2 & 79.5 & 66.9 & 0.0 & 0.0 & 0.0 & 58.0 & 0.0 & 0.0 & 0.0 & 80.3\\
			play & antmaze-med & 0.0 & 35.3$~\pm~$1.3 & 40.7$~\pm~$14.7 & 0.0 & 0.0 & 0.0 & 0.0 & 0.0 & 1.7 & 0.0 & 0.0 & 0.0 & 0.0\\
			play & antmaze-large & 0.0 & 20.2$~\pm~$14.8 & 23.6$~\pm~$14.7 & 0.0 & 0.0 & 0.0 & 0.0 & 0.0 & 2.2 & 0.0 & 0.0 & 0.0 & 0.0\\
			diverse & antmaze-umaze & 47.0 & 61.2$~\pm~$3.3& 63.6$~\pm~$9.5 & 48.0 & 6.3 & 37.7 & 0.0 & 0.0 & 52.0 & 0.0 & 0.0 & 0.0 & 57.3\\
			diverse & antmaze-med & 0.0 & 27.3$~\pm~$3.9 & 20.9$~\pm~$4.6 & 0.0 & 61.5 & 0.0 & 0.0 & 0.0 & 2.0 & 0.0 & 0.0 & 0.0 & 0.0\\
			diverse & antmaze-large & 0.0 & 41.2$~\pm~$4.2 & 31.4$~\pm~$12.1 & 0.0 & 0.0 & 0.0 & 0.0 & 0.0 & 1.0 & 0.0 & 0.0 & 0.0 & 0.0\\ \midrule
			random & mujoco-walker2d & 36.0 & 5.0$~\pm~$0.1 & 7.5$~\pm~$20.0 & 1.1 & 2.4 & 11.1 & 0.0 & 0.0 & 2.4 & 4.2 & 37.3 & -0.0 & 7.0\\
			random & mujoco-hopper & 0.9 & 8.0$~\pm~$0.8 & 30.2$~\pm~$4.4& 7.9 & 6.7 & 31.4 & 10.0 & 7.1 & 8.0 & 4.1 & 53.6 & 31.7 & 17.9\\
			random & mujoco-halfcheetah & 2.2 & 2.3$~\pm~$0.0 & 24.5$~\pm~$0.7 & 2.3 & 19.6 & 2.3 & 2.2 & 2.2 & 2.2 & 35.4 & 25.6 & 29.1 & 38.8\\
			med & mujoco-walker2d & 41.7 & 72.4$~\pm~$0.7 & 84.3$~\pm~$1.1 & 45.2 & 83.8 & -0.3 & 0.1 & -0.3 & 68.8 & -0.2 & 77.8 & 85.3 & 75.5\\
			med & mujoco-hopper & 40.0 & 50.0$~\pm~$5.7& 72.3$~\pm~$17.5 & 53.4 & 66.6 & 50.8 & 1.0 & 0.7 & 57.5 & 48.0 & 95.4 & 11.2 & 94.9\\
			med & mujoco-halfcheetah & 39.2 & 46.2$~\pm~$0.3 & 49.1$~\pm~$0.8 & 47.1 & 49.0 & 43.1 & 0.9 & 1.7 & 43.3 & 69.5 & 42.1 & 67.4 & 54.2\\
			med-replay & mujoco-walker2d & 2.2 & 49.5$~\pm~$1.6 & 109.7$~\pm~$9.8 & 41.8 & 88.2 & -2.4 & 1.7 & 4.3 & 51.4 & 69.4 & 49.8 & 83.8 & 56.0\\
			med-replay & mujoco-hopper & 8.1 & 39.1$~\pm~$2.2 & 98.1$~\pm~$2.0 & 53.2 & 97.0 & 30.6 & 1.2 & 6.1 & 54.7 & 39.1 & 93.6 & 5.7 & 73.1\\
			med-replay & mujoco-halfcheetah & 25.6 & 43.3$~\pm~$0.3 & 46.4$~\pm~$0.7 & 41.1 & 47.1 & 35.7 & 2.2 & 2.2 & 38.9 & 68.2 & 40.2 & 65.5 & 55.1\\
			med-expert & mujoco-walker2d & 73.4 & 110.7$~\pm~$0.2 & 109.7$~\pm~$9.8 & 110.7 & 109.4 & -0.2 & -0.1 & 1.1 & 75.7 & -0.3 & 95.6 & 111.2 & 96.1\\
			med-expert & mujoco-hopper & 36.0 & 77.2$~\pm~$0.6 & 111.4$~\pm~$0.4 & 76.4 & 106.8 & 40.5  & 7.9 & 0.7 & 106.4 & 3.3 & 108.7 & 69.8 & 111.1\\
			med-expert & mujoco-halfcheetah & 39.7 & 89.5$~\pm~$1.0 & 86.8$~\pm~$19.7& 91.0 & 90.8 & 50.6 & 2.2 & 2.3 & 92.6 & 72.7 & 95.6 & 77.5 & 90.0\\ \bottomrule
		\end{tabular}
	\end{center}
\end{sidewaystable}

\newpage

\section{Comparison between ROMI-CQL and FOMI-CQL}\label{appendix_fomi_cql}
Similar to Table~\ref{tab:abla_maze}, we compare ROMI-CQL with FOMI-CQL on the \emph{gym} tasks. From the results shown in Table~\ref{tab:abla_mujoco}, we can see that reverse imagination consistently outperforms forward imagination and that forward rollout sometimes even hurts the policy learning (e.g., in \emph{hopper-medium}). These empirical results are consistent with those on \emph{maze2d} and \emph{antmaze}. Together with Table~\ref{tab:mujoco} we find that ROMI-CQL does not achieve the best performance in \emph{halfcheetah}, which may be due to large model error.

\begin{table}[H]
	\centering
	\caption{Performance of ROMI-CQL and FOMI-CQL on Gym-MuJoCo tasks.}
	\begin{tabular}{l|l|c|c}
		\toprule
		Dataset type & Environment &  ROMI-CQL (ours) & FOMI-CQL \\
		\midrule
		random & walker2d & 7.5$~\pm~$20.0 & 2.1$~\pm~$0.5 \\
		random & hopper & 30.2$~\pm~$4.4 & 8.5$~\pm~$0.3 \\
		random & halfcheetah & 24.5$~\pm~$0.7 & 23.0$~\pm~$0.8 \\
		medium & walker2d & 84.3$~\pm~$1.1 & 84.2$~\pm~$2.0 \\
		medium & hopper & 72.3$~\pm~$17.5 & 1.9$~\pm~$0.1 \\
		medium & halfcheetah & 49.1$~\pm~$0.8 & 48.4$~\pm~$1.1 \\
		medium-replay & walker2d & 109.7$~\pm~$9.8 & 83.7$~\pm~$4.1 \\
		medium-replay & hopper & 98.1$~\pm~$2.6 & 94.2$~\pm~$4.4 \\
		medium-replay & halfcheetah & 47.0$~\pm~$0.7 & 46.6$~\pm~$0.4 \\
		medium-expert & walker2d & 109.7$~\pm~$5.3 & 109.6$~\pm~$0.3 \\
		medium-expert & hopper & 111.4$~\pm~$5.6 & 111.4$~\pm~$1.2 \\
		medium-expert & halfcheetah & 86.8$~\pm~$19.7 & 61.2$~\pm~$44.4 \\
		\bottomrule
	\end{tabular}
	\label{tab:abla_mujoco}
\end{table}

\section{Average Trajectory Discrepancy for \emph{maze2d-medium-sparse} and \emph{maze2d-large-sparse} }\label{appendix_discrepancy}

\begin{figure*}[h]
    \centering
        {\includegraphics[width=0.45 \textwidth]{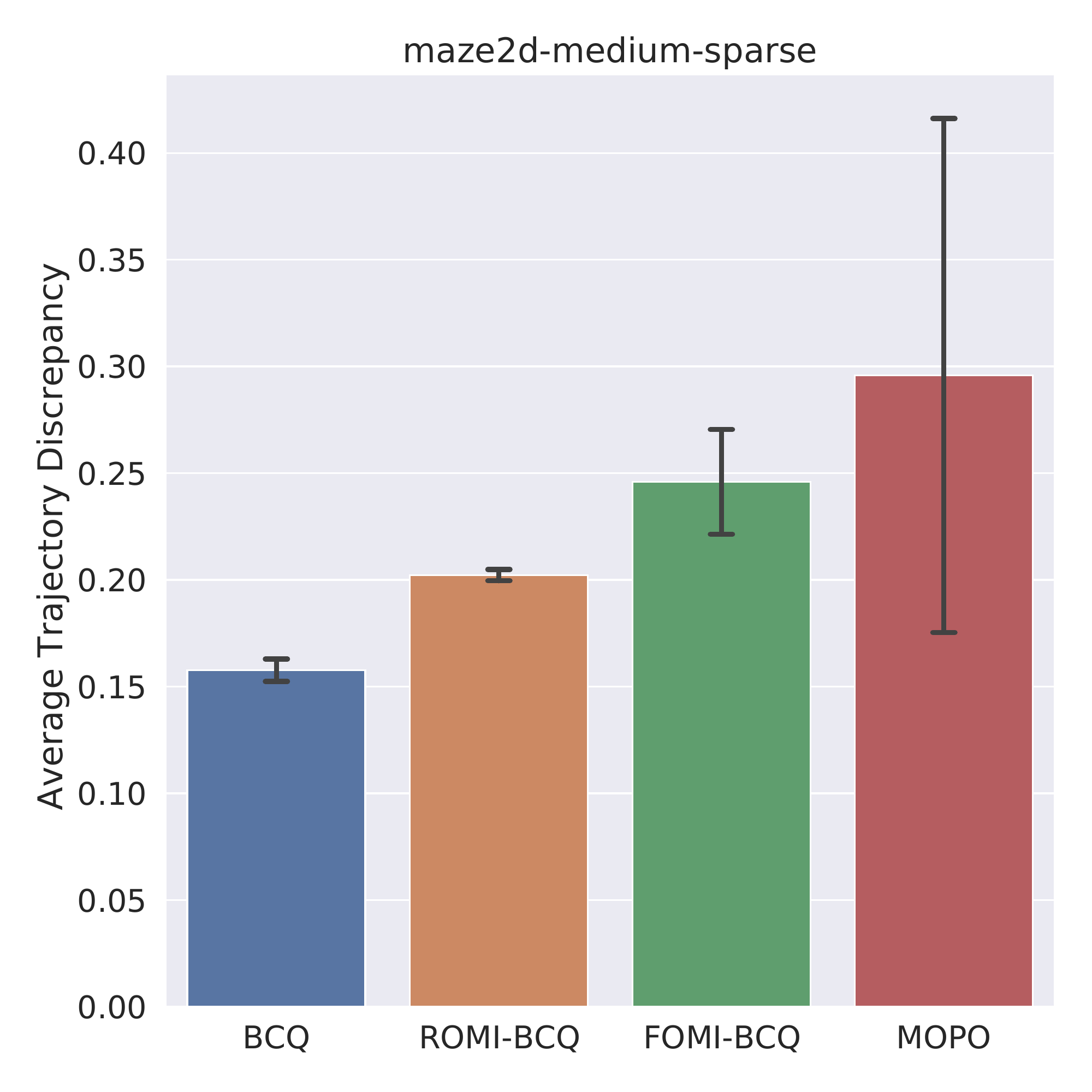}}
        {\includegraphics[width=0.45 \textwidth]{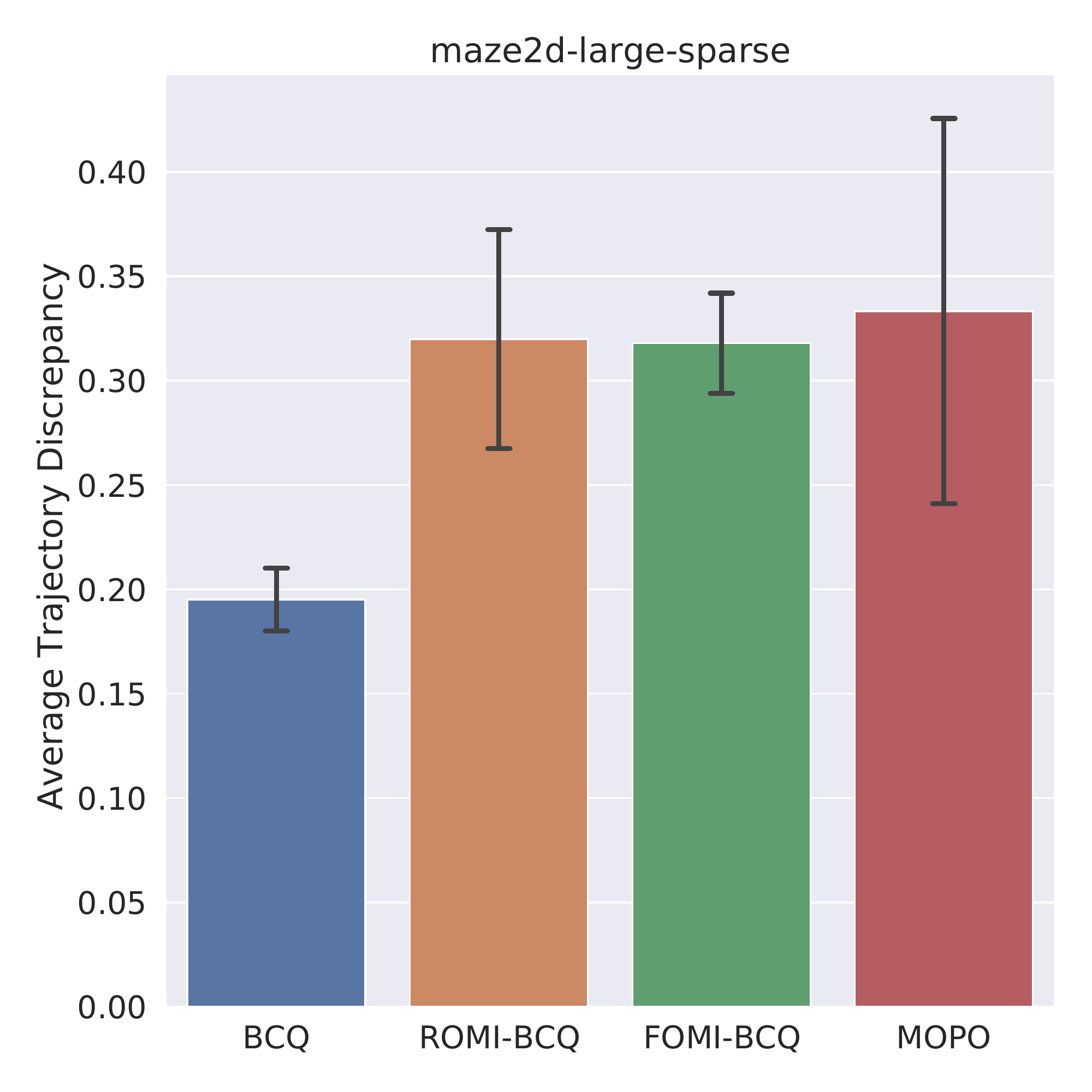}}
    \caption{Average Trajectory Discrepancy for \emph{maze2d-medium-sparse} and \emph{maze2d-large-sparse}.
    % \av{larger fonts}\zz{sort legends by rmse}
    }
    \label{fig:atd-comparison}
\end{figure*}

We show the average trajectory discrepancy defined by Eq.~(\ref{eq:atd}) in \emph{maze2d-medium-sparse} and \emph{maze2d-large-sparse} environments in Figure~\ref{fig:atd-comparison}. The result in \emph{maze2d-medium-sparse} is consistent with that in \emph{maze2d-umaze-sparse}, showing that the reverse imagination is more conservative than the forward counterpart. Note that the discrepancy of ROMI-BCQ is comparable with FOMI-BCQ and MOPO in \emph{maze2d-large-sparse} environment, as the maze is with relatively narrow paths to the goal.

\section{Ablations on Rollout Length}
To investigate the effect of the rollout length more clearly, we conduct an ablation study by varying this hyperparameter on the \emph{maze2d} and \emph{antmaze} environments. The results are listed in Table~\ref{tab:rollout_maze2d} and~\ref{tab:rollout_antmaze}. We find that on \emph{maze2d}, ROMI performs equally well or even better when increasing the rollout length to 20, because the model is accurate enough for longer horizon imagination. However, increasing ROMI's rollout length has a negative impact on \emph{antmaze}, where model cannot predict future states accurately after multi-step rollout.

\begin{table}[H]
	\caption{Ablations on rollout length on \emph{maze2d}. * indicates the hyperparameter used in the main paper.}
	\label{tab:rollout_maze2d}
	\centering
	\begin{tabular}{l|c|c|c|c}
		\toprule
		Environment & len=1 & len=5* & len=10 & len=20 \\
		\midrule
		sparse-maze2d-umaze & 82.0$~\pm~$31.5 & 139.5$~\pm~$3.6 & 134.4$~\pm~$17.2 & 135.6$~\pm~$7.3 \\
		sparse-maze2d-medium & 70.4$~\pm~$28.6 & 82.4$~\pm~$15.2 & 82.9$~\pm~$6.2 & 104.5$~\pm~$54.7 \\
		sparse-maze2d-large & 120.3$~\pm~$18.3 & 83.1$~\pm~$22.1 & 116.2$~\pm~$49.3 & 76.8$~\pm~$17.2 \\
		dense-maze2d-umaze & 68.0$~\pm~$11.5 & 98.3$~\pm~$2.5 & 113.6$~\pm~$7.0 & 99.2$~\pm~$10.3 \\ 
		dense-maze2d-medium & 88.5$~\pm~$20.9 & 102.6$~\pm~$32.4 & 93.2$~\pm~$16.3 & 83.0$~\pm~$10.1 \\
		dense-maze2d-large & 120.2$~\pm~$4.8 & 124.0$~\pm~$1.3 & 113.6$~\pm~$17.9 & 97.5$~\pm~$40.7 \\
		\bottomrule
	\end{tabular}
\end{table}

\begin{table}[H]
	\caption{Ablations on rollout length on \emph{antmaze}. * indicates the hyperparameter used in the main paper.}
	\label{tab:rollout_antmaze}
	\centering
	\begin{tabular}{l|c|c|c|c}
		\toprule
		Environment & len=1 & len=5* & len=7 & len=10 \\
		\midrule
		fixed-antmaze-umaze & 69.7$~\pm~$6.5 & 68.7$~\pm~$2.7 & 44.2$~\pm~$29.6 & 54.8$~\pm~$25.4 \\ 
		play-antmaze-medium & 36.8$~\pm~$6.8 & 35.3$~\pm~$1.3 & 33.5$~\pm~$14.9 & 28.0$~\pm~$10.4 \\
		play-antmaze-large & 37.5$~\pm~$6.1 & 20.2$~\pm~$14.8 & 7.3$~\pm~$12.7 & 0 \\
		diverse-antmaze-umaze & 12.8$~\pm~$19.7 & 61.2$~\pm~$3.3 & 58.3$~\pm~$18.5 & 42.8$~\pm~$14.3 \\ 
		diverse-antmaze-medium & 25.3$~\pm~$4.9 & 27.3$~\pm~$3.9 & 18.2$~\pm~$9.3 & 13.5$~\pm~$4.6 \\
		diverse-antmaze-large & 31.2$~\pm~$2.4 & 41.2$~\pm~$4.2 & 32.8$~\pm~$5.1 & 25.0$~\pm~$5.8 \\
		\bottomrule
	\end{tabular}
\end{table}